\UseRawInputEncoding

\documentclass[APA,STIX2COL]{WileyNJD-v2}

\usepackage{balance}
\usepackage{graphicx}
\usepackage{subcaption}

\usepackage{threeparttable,booktabs}

\makeatletter
\def\NAT@aysep{,}
\makeatother

\articletype{Original Article}%

\received{20 Dec. 2021}
\revised{}
\accepted{}

\begin{document}

\title{A comparative study of attention mechanism and generative adversarial network in facade damage segmentation}

\author[1,3]{Fangzheng Lin*}
\author[1]{Jiesheng Yang}
\author[2]{Jiangpeng Shu}
\author[1]{Raimar J. Scherer}

\authormark{LIN \textsc{et al}}

\address[1]{\orgdiv{Institute of Construction Informatics}, \orgname{Dresden University of Technology}, \orgaddress{\state{Dresden}, \country{Germany}}}
\address[2]{\orgdiv{Collage of Civil Engineering and Architecture}, \orgname{Zhejiang University}, \orgaddress{\state{Hanhgzhou}, \country{China}}}
\address[3]{\orgdiv{Deep Learning Center}, \orgname{Changzhou Microintelligence Co., Ltd.}, \orgaddress{\state{Changzhou}, \country{China}}}

\corres{Fangzheng Lin, Ph.D. Candidate, Institute of Construction Informatics, Technische Universität Dresden, Dresden 01069, Germany.  \linebreak
\email{fangzheng.lin@tu-dresden.de}}


\abstract[Abstract]{
 
Semantic segmentation profits from deep learning and has shown its possibilities in handling the graphical data from the on-site inspection. As a result, visual damage in the facade images should be detected. Attention mechanism and generative adversarial networks are two of the most popular strategies to improve the quality of semantic segmentation. With specific focuses on these two strategies, this paper adopts U-net, a representative convolutional neural network, as the primary network and presents a comparative study in two steps. First, cell images are utilized to respectively determine the most effective networks among the U-nets with attention mechanism or generative adversarial networks. Subsequently, selected networks from the first test and their combination are applied for facade damage segmentation to investigate the performances of these networks. Besides, the combined effect of the attention mechanism and the generative adversarial network is discovered and discussed.

}


\jnlcitation{\cname{%
\author{Lin F.}, 
\author{Yang J.}, 
\author{Shu J.}, 
\author{Scherer R. J.},
\author{Guo J.},}, 
\ctitle{A comparative study of attention mechanism and generative adversarial network in facade damage segmentation}. 
\cjournal{Comput Aided Civ Inf.}, \cvol{20xx;xx:x--x}.}

\maketitle

\footnotetext{\textbf{Abbreviations:} CV, Computer Version; AEC, Architecture, Engineering and Construction; GAN, Generative Adversarial Network; CNN, Convolutional Neural Network; ANN, Artificial Neural Networks; FCN, Fully Convolution Network}

\section{Introduction}\label{sec1}

The deep learning based computer version (CV) concept implies the ultimate expectation of technology development, i.e., to completely replace or even exceed the human eyes' functionality. Nowadays, state-of-the-art is still far from the goal yet, but CV technology develops by handling tasks step by step. The current CV technology can bring advantages in processing complex images and massive graphical data. Generally, CV usage can be categorized into four categories: classification object detection, semantic segmentation, and instance segmentation. The detected object can be rendered in a specific color in the prediction results from semantic segmentation. This kind of labeling presents more precise information of objects' sizes positions than object detection. Instance segmentation is a more advanced application to detect each object instance in a photo. 

Due to the outstanding effects of CV, the Architecture, Engineering, and Construction (AEC) industry explores the possibilities of the related applications, especially in the field of Structural Health Monitoring. The visible damage in buildings and constructions is supposed to be detected, recognized, and classified by means of the deep learning based CV. 
 
As the first step, on-site inspection is executed to collect a structure's one-hand real-time data. Hence, visual inspection, a conventional, laborious but practical approach, is unavoidable. Nevertheless, the manual inspection approach still has to face several obstacles. The amount of manual data collection could be restricted. The inspectors' life might be under threat, for instance, if they step into a building damaged in an earthquake or they stand under a sea-crossing bridge. Broader utilization of drones breaks through the limitation of humans and yields an efficient approach to damage data acquisition. A sufficient amount of image data can be acquired from on-site inspection using drones. This convenient and efficient approach improves the productivity of data collection productivity and changes the conventional processing methods of inspection results. 

This paper presents a comparative study on the U-net based semantic segmentation using attention mechanism and Generative Adversarial Networks (GANs). In section \ref{sec2}, literature is collected and reviewed to clarify the state-of-the-art Convolutional Neural Network (CNNs), semantic segmentation, and the corresponding application in the AEC industry. The newly proposed combinations of U-net and attention mechanism are elaborated in section \ref{sec3}. Afterward, GANs are employed to enhance the semantic segmentation, and the proposed discriminators are depicted in section \ref{sec4}. In section \ref{sec5}, a lightweight test is executed to select the CNNs with optimal performances, and the selected CNNs are applied for facade damage segmentation. Section \ref{sec6} specifically discussed the combined effect of attention mechanism and GANs. Finally, section \ref{sec7} concludes the entire research work, and suggests the next research steps in the futures.

\section{State of the art}\label{sec2}

As a kind of modern computer science technology, Deep Learning is directly subject to the rapid development of computing capability. The corn of deep learning lies in the intensive training of Artificial Neural Networks (ANNs). Compared with other derived ANN, e.g., Recurrent Neural Networks and novel Graph Convolutional Networks, CNN are distinguished by their strength in image processing. Initial functions i.e. recognition and classification, cannot fulfill the requirement from the industry. Advanced approaches and the corresponding application are proposed, studied and implemented. This section retrospects the growth of CNN and particularly reviewed the related work in segmentation.

\subsection{The convolutional networks}

Autonomous vehicles have become one of the most famous AI products in recent years. Its development is directly subject to one vital technology, computer version, which experienced a development journal in the past years. The first work on modern CNNs occurred in the 1990s, inspired by the recognition. \cite{lecun_gradient-based_1998} demonstrated that a CNN model which aggregates more specific features into progressively more complicated features could be successfully used for handwritten character recognition. MNIST \citep{lecun_mnist_1998} is a now-famous dataset that includes images of handwritten digits paired with their true label of 0, 1, 2, 3, 4, 5, 6, 7, 8, or 9. A CNN model is trained on MNIST by giving it an example image, asking it to predict what digit is shown in the image, and then updating the model's settings based on whether it predicted the digit identity correctly or not. 

Around 2012 the CNN enjoyed a massive surge in popularity after \cite{krizhevsky_imagenet_2017} published AlexNet and achieved state-of-the-art performance labeling pictures in the ImageNet challenge. Similar to MNIST, ImageNet \citep{stanford_vision_lab_imagenet_2020} is a public, freely available dataset of images and corresponding true labels. However, instead of focusing on handwritten digits labeled 0 – 9, ImageNet focuses on ``natural images'' or pictures of the world, labeled with a variety of descriptors including ``amphibian,'' ``furniture,'' and ``person.'' \cite{simonyan_very_2015} trained their proposed CNN VGG-16 for weeks and achieved top 5 accuracies in the ImageNet test. The massive data in the ImageNet were manually labeled and prepared. However, the laborious work guarantees the comprehensiveness and diversity of this dataset. Thus, it keeps encouraging innovative work of CNNs and also validates them. \cite{he_deep_2015} conceived the residual learning and constructed the ResNet, which is much deeper than VGG but has a lower complexity. 

All the CNNs mentioned above were initially developed as classifiers. Nevertheless, they have become the representative foundation of more advanced CV technology, e.g., object detection, semantic segmentation, and instance segmentation. The popular Fast R-CNN \citep{girshick_fast_2015} took the VGG-16 as the backbone for object detection. \cite{ou_moving_2019} extended the ResNet-18 with an encoder-decoder structure. The developed method is capable of segmenting objects from dynamic scenes. Instance segmentation requires both object detection and semantic segmentation. \cite{zhao_deep_2021} adopted ResNet-101 to classify leakage-areas of shield tunnel linings and utilized the Mask R-CNN \citep{he_mask_2018} to label these different leakage-areas.   

\subsection{Semantic segmentation}

An image is nothing but a collection of pixels. Image segmentation is the process of classifying each pixel in an image belonging to a particular class and hence can be thought of as a classification problem per pixel. There are two kinds of image segmentation, namely semantic segmentation, and instance segmentation. The former is the process of classifying each pixel belonging to a particular label. It does not distinguish different instances of the same object. Instance segmentation differs from semantic segmentation in the sense that it gives a unique label to every instance of a particular object in the image. 

Semantic segmentation is a classic topic, which has been studied for decades. Dating back to 1992, \cite{beek_semantic_1992} faced the head-and-shoulder scene in video telephony and developed a segmentation system to locate the speaker, the face, and the eyes automatically and sequentially in a video. Conventional approaches and mathematical modes were adopted in that system, such as model-based coding and fuzzy membership functions. This paper, which reveals the state of semantic segmentation at the early stage in a certain aspect.  is a representative but non-ephemeral one. \cite{csurka_efficient_2011} inherited that thought and conceived an efficient but straightforward approach enhanced by the Fischer kernel. Based on open-source datasets, the results provided a comparable performance on semantic segmentation, for which CNNs are nowadays prevailing deducted.

\subsection{Fully Convolution Networks and the U-net}

Numerous pieces of research work, which present diverse convolutional networks, endorse that CNNs improve the quality of semantic segmentation significantly with respect to efficiency and accuracy. In 2014, \cite{long_fully_2015} published the Fully Convolution Networks (FCNs). They interchanged the fully connected layers of a CNN classifier. e.g., LeNet \citep{lecun_gradient-based_1998}, into convolution layers. In experiments, FCN was tested and delivered satisfactory prediction results. FCN is a concept rather than a specific permanent convolutional net. It can be implemented with the existing CNNs that serve as the backbone. \cite{mou_relation-augmented_2019} adopted the FCN, took the VGG-16 \citep{simonyan_very_2015} net as the backbone, and inserted the relation modules for semantic segmentation in aerial scenes. \cite{tong_evidential_2021} composed an FCN and a Dempster-Shafer layer so that the pixel-wise feature maps are extra extracted from an input image to promote the prediction quality. FCN might be one of the most popular CNNs since its appearance. A large number of papers contributed to enhancing FCNs in various ways. However, most of them do not think outside the box of improving FCNs by adding a component. 

Beyond that, novel architectures are proposed to explore other approaches to semantic segmentation. U-net \citep{ronneberger_u-net_2015} is a typical one; it is established still on the basis of the encoding-decoding principle.  Nevertheless, utilizing the skip connection to leverage the output from down-sampling can be interpreted as an innovative strategy in the U-net in order to reduce the required training data amount. Since an original CNN is usually simply designed. Thus there is much potential for further development. With focuses on the functionality of skip connections, U-net++ \citep{zhou_unet_2018} and U-net 3+ \citep{huang_unet_2020} share a similar idea of enhancement. In the U-net++, the encoder and decoder sub-networks are nests to replace the original skip connections. This network can be interpreted as a more dense and complicated version of the U-net. \cite{huang_unet_2020} proposed the U-net 3+, which abandons the nested skip connections and adopts full-scale skip connections. By means of that, the output of an encoding layer is transmitted to each sub-decoding layer. In addition, the labels are used in the decoding layers to adjust the loss calculation.  Experiments in both papers were executed with the state-of-the-art dataset, and the consequence is that these kinds of reinforcement positively influence semantic segmentation. Changing skip connection is one direction to obtaining a more powerful U-net. Another direction would be transformation or accumulation. W-net \citep{xia_w-net_2017} is not a pure concatenation of two U-nets; two specific loss functions maximize semantic segmentation. \cite{mehta_m-net_2017} added a left leg and a right one to a U-net named M-net. An effective semantic segmentation materializes through concatenating the initial input and the output from each encoding layer and concatenating the output of the bottom layer and the output from each decoding layer.

\subsection{Generative adversarial networks}

Since the generative adversarial network was introduced to the world by \cite{Goodfellow2014GenerativeAN}, its related developments and applications present a tendency of exploration. This design opens a new window of CNN application. In the definition, the training of a GAN should be proceeded by two steps in each epoch (see Eq. \ref{eqGAN}). When an epoch \(\mathbb{E}\)  starts, the discriminator \(D(x)\) is trained firstly with the generator's \(G(x)\) initial output and the labels \(z\). As long as \(D(x)\) reaches the maximum \(max_{D}\), the training of discriminator is completed. Subsequently, the trained discriminator feedbacks information to the generator, the generator's training is then launched by feeding the input data \(x\) until the generator's minimum \(min_{G}\) is obtained. This ingenious orchestration between a generator and a discriminator encourages researchers to mine its potential in semantic segmentation. 

\begin{equation}\label{eqGAN}
\begin{split}
\min_{G} \max_{D} V\left ( D,G \right ) & =\mathbb{E}_{x\sim P_{data}\left ( x \right )}\left [ logD\left ( x \right ) \right ] \\
&+ \mathbb{E}_{z\sim P_{z}\left ( z \right )}\left [ log\left ( 1 - D\left ( G\left ( z \right ) \right ) \right ) \right ]
\end{split}
\end{equation}

Various domains have benefited from the semantic segmentation using GANs. \cite{decourt_semi-supervised_2020} combined the ResNet-101 \citep{He2016DeepRL} as the generator and an FCN as the discriminator to build a GAN for MRI image segmentation. Object identification in medical images is indeed an issue that challenges researchers. Brain tumors in MRI pictures can be segmented through a pre-trained GAN \citep{ghassemi_deep_2020}.  \cite{zhang_unsupervised_2020} accomplished multi-organ segmentation for X-ray images. Apart from medical images, remote sensing pictures are often requested to be segmented. \cite{huang_object-level_2021} connected the U-net as the generator and an FCN as a discriminator to identify the ships in remote sensing images of a harbor. \cite{xiong_end--end_2020} proposed a more complicated GAN architecture to promote semantic segmentation. In this combination, the FCN segments the buildings, roads, and forests in the input remote sensing images, and the GAN is trained to optimize the first segmentation.

\subsection{Application in the AEC industry}

While computer scientists developed new CNN models or CV-based approaches for a generic and holistic application, the practitioners in the AEC industry adore optimizing these models or approaches to address specific issues. \cite{dais_automatic_2021} proposed a comprehensive assessment system for future landscapes. In this system, an open-source toolbox was adopted to segment the landscape images. The predictions were visualized with live videos in a mixed reality system for landscape assessment. In a single building project, semantic segmentation can be applied for building component classification. \cite{czerniawski_automated_2020} considered the features of color and depth in training a deep convolutional network for this objective. The authors also adopted two trained CNNs to label the object in point clouds of building interiors \citep{ma_semantic_2020}. Objects in infrastructures can also be identified via semantic segmentation. \cite{chen_image-based_2020} optimized the deep convolutional network to detect the weak interlayers in tunnel faces and evaluate them quantitatively. Compared with the FCN, the proposed network shows a higher accuracy and less training

Generally, the encoding-decoding principle denotes the essential thought behind semantic segmentation using CNNs. No matter FCNs, U-nets, or GANs, all of them obey this principle. Although copious distinct networks \citep{milletari_v-net_2016, zheng_rethinking_2021, shim_lightweight_2020} are conceived to address semantic segmentation, the deja-vu about the three networks always exists in these publications. In terms of the AEC industry, visible or superficial damage identification has become a specific vital domain, such as detecting cracks, spall, or corrosion. Some papers emphasize the networks themselves; meanwhile, some propose a workflow that encompasses the function of semantic segmentation and the strategy of data acquisition. For example, drones promote the efficiency and safety of taking images of the damage in, e.g., bridge super-/substructures \citep{seo_drone-enabled_2018} or building roofs \citep{inoguchi_establishment_2019}. The majority of related publications propose novel CNN architectures to compare it with other networks. Generally, only the calculated metrics are compared to prove the feasibility and advancement of the proposed network. However, few papers analyzed the relation between the metrics and the design of CNNs. This research work proposes novel CNNs and simultaneously investigates the effects from different combinations of networks through comparative study. 

\section{The U-net based architectures for semantic segmentation}\label{sec3}

In this section, the original U-net and the attention mechanism are initially manifested. The redistributions of attention mechanisms are studied. Inspired by the attention U-net \citep{schlemper_attention_2019}, two novel CNNs are proposed to upgrade the usage of input data during training and test. 

\begin{figure*}[htbp]
    \centering
    \begin{subfigure}[b]{0.3\textwidth}
        \centering
            \includegraphics[angle=90,origin=c,width=0.8\textwidth]{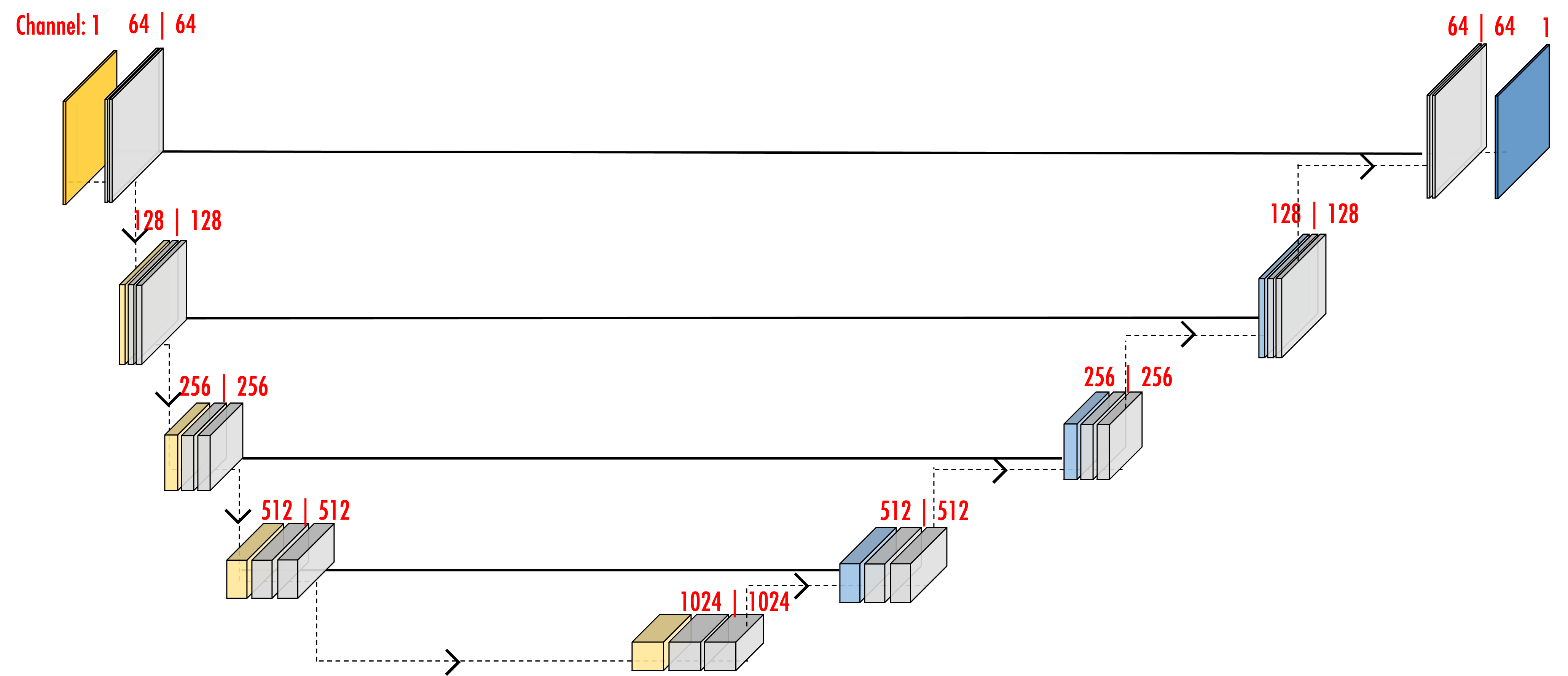}
        \caption{The U-net}
        \label{fig:attn1}
    \end{subfigure}
    \hfill
    \begin{subfigure}[b]{0.3\textwidth}
        \centering
            \includegraphics[angle=90,origin=c,width=0.8\textwidth]{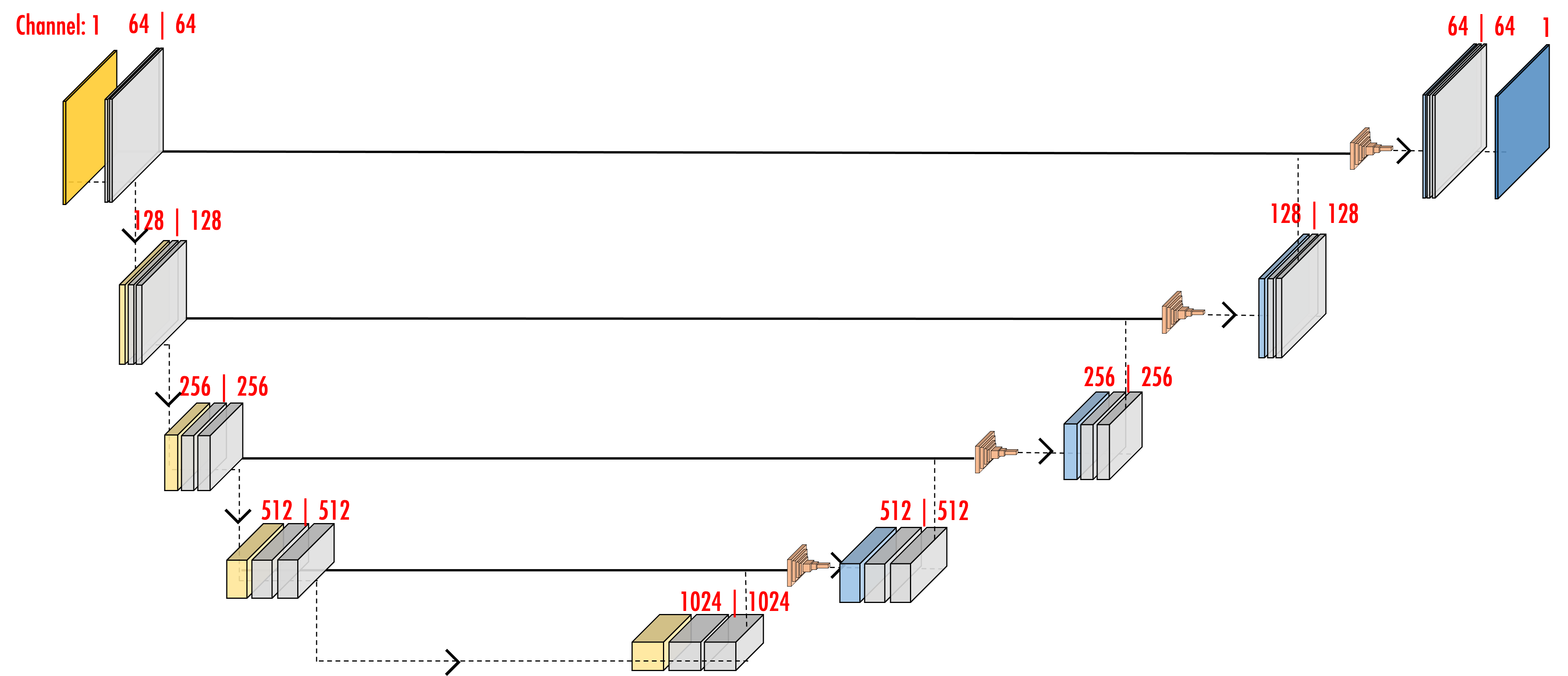}
        \caption{The attention U-net}
        \label{fig:attn2}
    \end{subfigure}
    \hfill
    \begin{subfigure}[b]{0.3\textwidth}
        \centering
            \includegraphics[angle=90,origin=c,width=0.8\textwidth]{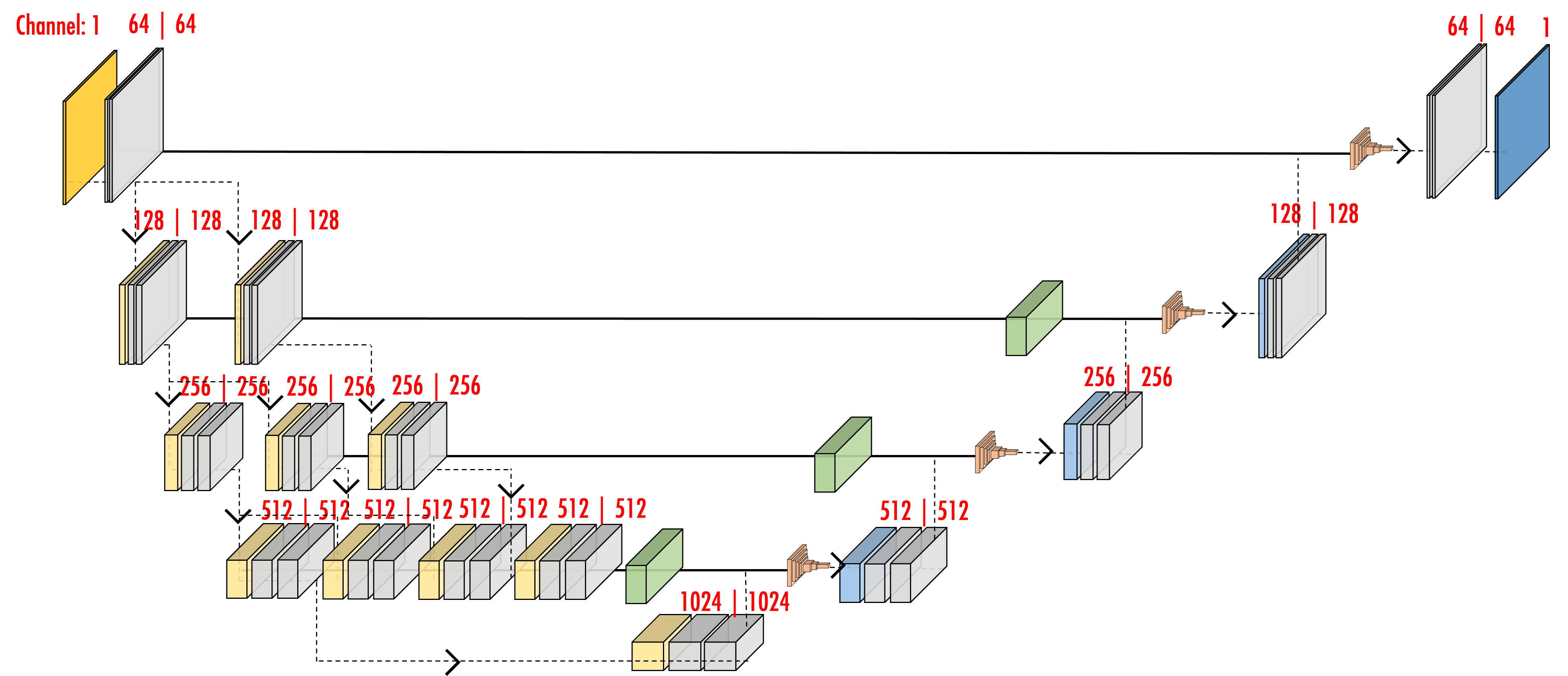}
        \caption{The advanced attention U-net}
        \label{fig:attn3}
    \end{subfigure}
    \\
    \begin{subfigure}[b]{\textwidth}
        \centering
            \includegraphics[angle=90,origin=c,width=0.7\textwidth]{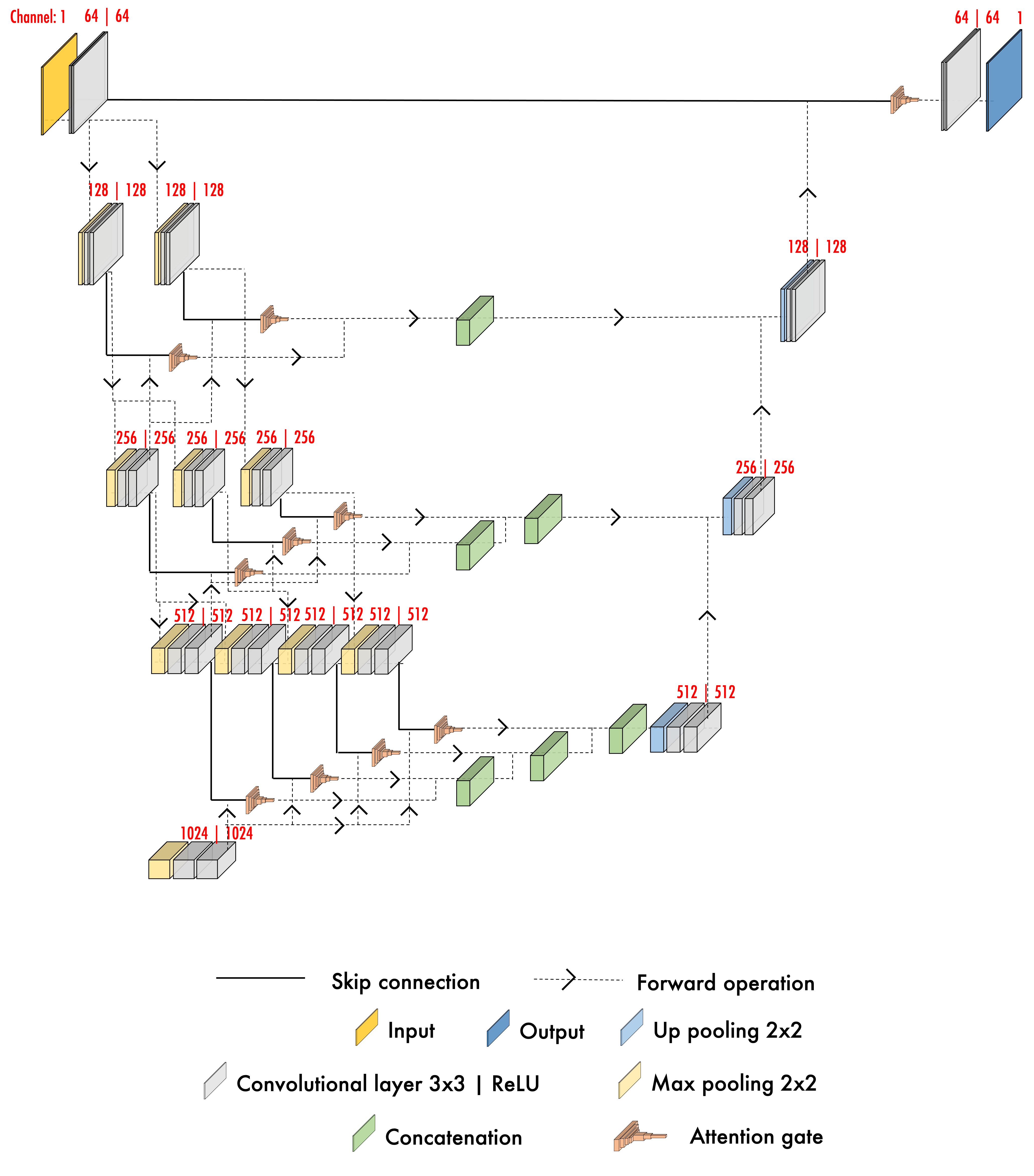}
        \caption{The full attention U-net}
        \label{fig:attn4}
    \end{subfigure}
    \caption{The U-net and the studied U-net based networks}
\end{figure*}
  
\subsection{U-net}

The U-net \citep{ronneberger_u-net_2015} is designed as a encoder-decoder structure with four layers of down- and upsampling. The default input image should have one channel. In the case of colorful inputs with three channels, the phase of data preparation can switch them into mono-colored images. The first layer of downsampling is composed of two convolution operations. The former convolution reshapes the input data, while outputs of the latter one remain the size of the inputs. Unlike the first layer in the encoder, a max-pooling stage is added to the other encoder layers. Along with the downsampling process, the defined input and output channels change. The bottom layer is treated as the extension of the encoder layer. Hence, it is built with the same structure and corresponding channel adaptation. Decoder layers are constructed strictly identically to the encoders apart from the pooling stage. Up pooling serves to concatenate the skip connection and the output of the downer layer and subsequently scale on the processed data. The softmax function is recommended by \cite{ronneberger_u-net_2015} to be applied in the last stage of the topside decoder layer. However, other functions having equivalent effects can be employed as well. The skip connection, the output from each encoder layer, is a feature, which distinguishes the U-net from other FCN based architectures. By using skip connections in the U-net, fewer images can be fed for training and achieve comparable semantic segmentation results.

\subsection{Attention mechanism}

Psychology defines attention as the cognitive process of selectively focusing on a few objects while disregarding others. A neural network is regarded as a reduced effort to imitate human brain activities. Attention Mechanism is also an effort to accomplish the same action by concentrating selectively on certain essential items while disregarding other things in deep neural networks.

\cite{bahdanau_neural_2015} implemented the attention mechanism in an RNN for neural machine translation. It was the very first appearance of an attention mechanism in artificial neural networks. The RNNs, oriented to language translation, are usually developed in the encoder-decoder. Generally, updating translation quality entails a relatively complicated RNN, which means multiple layers and a more extended structure. Consequently, the vital features in the input data could be reduced through convolutional layers in the decoder. Therefore, the attention mechanism, which was applied initially to address the issue, is dedicated to improving the possibility of the vital feature extraction in the decoder. For example, \cite{usama_attention-based_2020} combined an RNN and a CNN for the sentiment classification of short text. As the RNN works as the encoder, the attention within assists in focusing on essential text features. 

The performance of the attention mechanism in RNNs inspires researchers to explore its potential in CNNs. \cite{tian_attention-guided_2020} inserted the attention mechanism in a CNN to denoise the images automatically. In that CNN, the attention block exploits the two steps to implement the attention mechanism. The first step uses convolution of size 1 $\times$ 1 to compress the obtained features into a vector as the weights for adjusting the previous stage. The second step utilizes obtained weights to multiply the previous layer's output for extracting more prominent noise features. \cite{hang_hyperspectral_2021} employed attention mechanism in image detection. In the frame of that research, a dual-CNN group was proposed. A spectral attention module and a spatial one, integrated into the CNN group respectively, tried to extract the features from different dimensions. The former focuses the input along the direction of channels, and the latter evokes the feature vertically to the channels. The experiment verified the positive effects of two novel attention modules' collaboration in image classification. 

\begin{figure}[htbp]
    \centering
        \includegraphics[width=0.5\textwidth]{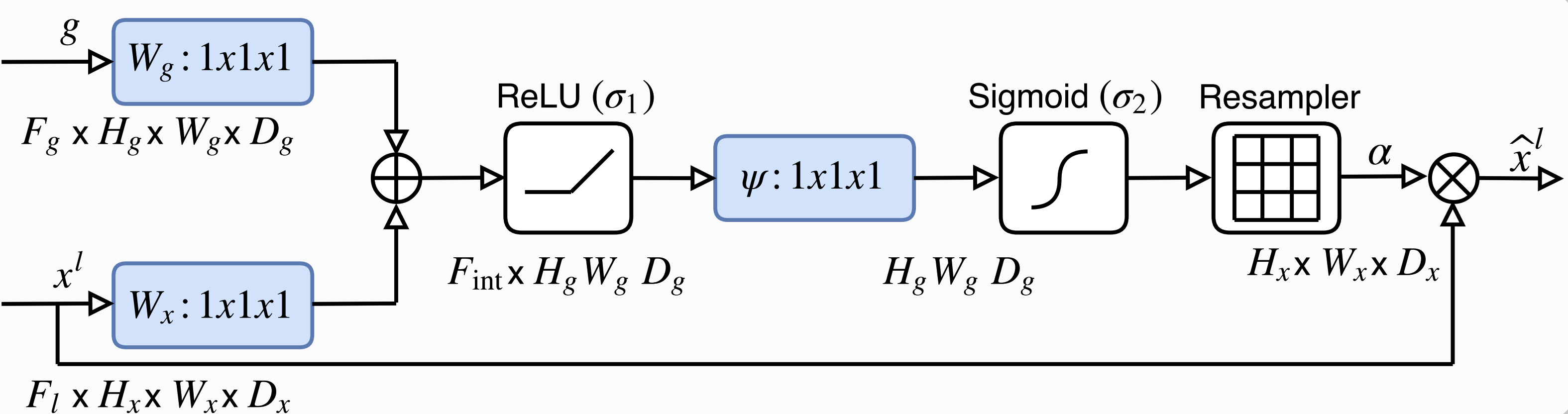}
    \caption{The diagram of the attention gate \citep{schlemper_attention_2019}}
    \label{fig:attenGate}
\end{figure}

The attention gate \citep{schlemper_attention_2019} is constructed with more stages, as Fig. \ref{fig:attenGate} illustrates. The previous attention implementations combined the data and that after attention processing once. On the contrary, the attention gate leverages skip connections, the unique feature of a U-net, named the attention U-net in Fig. \ref{fig:attn2}. The output from the previous stage is firstly concatenated with a skip connection and subsequently serves as the input. After the attention processing, the input is combined with the outputs and exported out of the attention gate. The corresponding upgrade of prediction quality was validated in the experiment of medical image segmentation. The attention mechanism is merely an abstract concept; hence, it can be implemented in various forms based on different usages. A practical attention module has not to be that sophisticated. The vital issue lies in feature extraction. The more features can be filtered out from the inputs, the better prediction results can be obtained. 

To the author's best knowledge, \cite{schlemper_attention_2019} contributed firstly to the combination of the U-net and attention mechanism. Attention gates inherit the conception of the attention mechanism and adjust it for the U-net through creating an interface to the skip connections. Attention gates are located before the concatenation operation in each decoding layer. The value $g$ denotes the skip connection produced by the encoder layer on the same level, and the value $x^l$ represents the upsampled data from the lower decoder layer. Though these two sets of data, which actually are two tensors, present with different sizes. The output of the attention gate should have the same size of $g$ before the concatenation with $x^l$. The idea behind attention gets leis that the upsampled data is reused to upgrade the prediction of the possible image regions of interest, which is merely realized by the skip connection in the U-net.    

\subsection{The advanced attention U-net}

\cite{schlemper_attention_2019} focused mainly on the decoder in the attention U-net. However, the encoder was not touched. Since the attention U-net proves that the accumulation of the outputs from encoders is capable of facilitating the U-net's performance in semantic segmentation, it is reasonable to expect even better results from a more comprehensive accumulation of encoders' outputs. The advanced attention U-net (see Fig. \ref{fig:attenGate}) is to constructed to leverage the data in skip connections. 

\cite{zhou_unet_2018} noticed this issue as well and proposed the U-net++ with more en-/decoder layers and sophisticated skip connection network between the layers. \cite{huang_unet_2020} proposed the U-net 3+, in which each decoder layer adopts the outputs from encoder layers on the same or lower levels, and related decoder layers also apply the output of the bottom layer. Aiming at a higher accuracy in semantic segmentation, the U-net 3+ combined several loss functions to present a correct gradient descent. Analog to these two networks, the advanced attention U-net proposed in this paper emphasizes the inherent structure of the U-net but is built as a novel architecture to find a compromise solution. By means of that, the output of an encoder layer is convoluted and transmitted to all the lower layers. For example, the topside output is operated three times so that the singly or repetitively convoluted output of the topside layer stands in each layer beneath. All the outputs on one level are concatenated as a tensor, which moves to the attention gate as the parameter $g$ in Fig. \ref{fig:attenGate}. The advanced attention U-net is designed to adopt typical loss functions, and no novel loss function has to be assembled or invented.  

\subsection{The full attention U-net}

While the advanced attention U-net obeys the order concatenation-to-attention, the full attention U-net is designed to explore a reverse way for the comparative study. Deriving from the advanced attention U-net, the full attention U-net switches the order of concatenation and attention. Except for the topside layer, encoded data on a level and one on the adjacent lower level are transported into an attention gate respectively as the $g$ and $x^l$ in Fig. \ref{fig:attn4}. Afterward, all the tensors produced by attention gates are concatenated as one tensor for upsampling in the decoder. The study of the order of concatenation and attention is barely to be seen in the related publications. Therefore, the comparative study in the paper is expected to investigate the effects caused by certain operations in two completely reverse orders.

\section{Enhancement with Generative Adversarial Networks}\label{sec4}

This section elaborates on the solution to enhance the U-net utilizing GANs. Therefore, in the frame of GANs, the aforementioned CNNs are to be adopted as the generator. And the discriminators are supposed to be studied. Thus, in order to investigate the segmentation capability of discriminators, four discriminators are proposed with the consideration of dual-dimensions of complexity, namely the layer number (length) of a discriminator and the stage number (thickness) of a single layer. 

\begin{figure*}[t]
     \centering
     \begin{subfigure}[b]{0.49\textwidth}
         \centering
            \includegraphics[width=\textwidth]{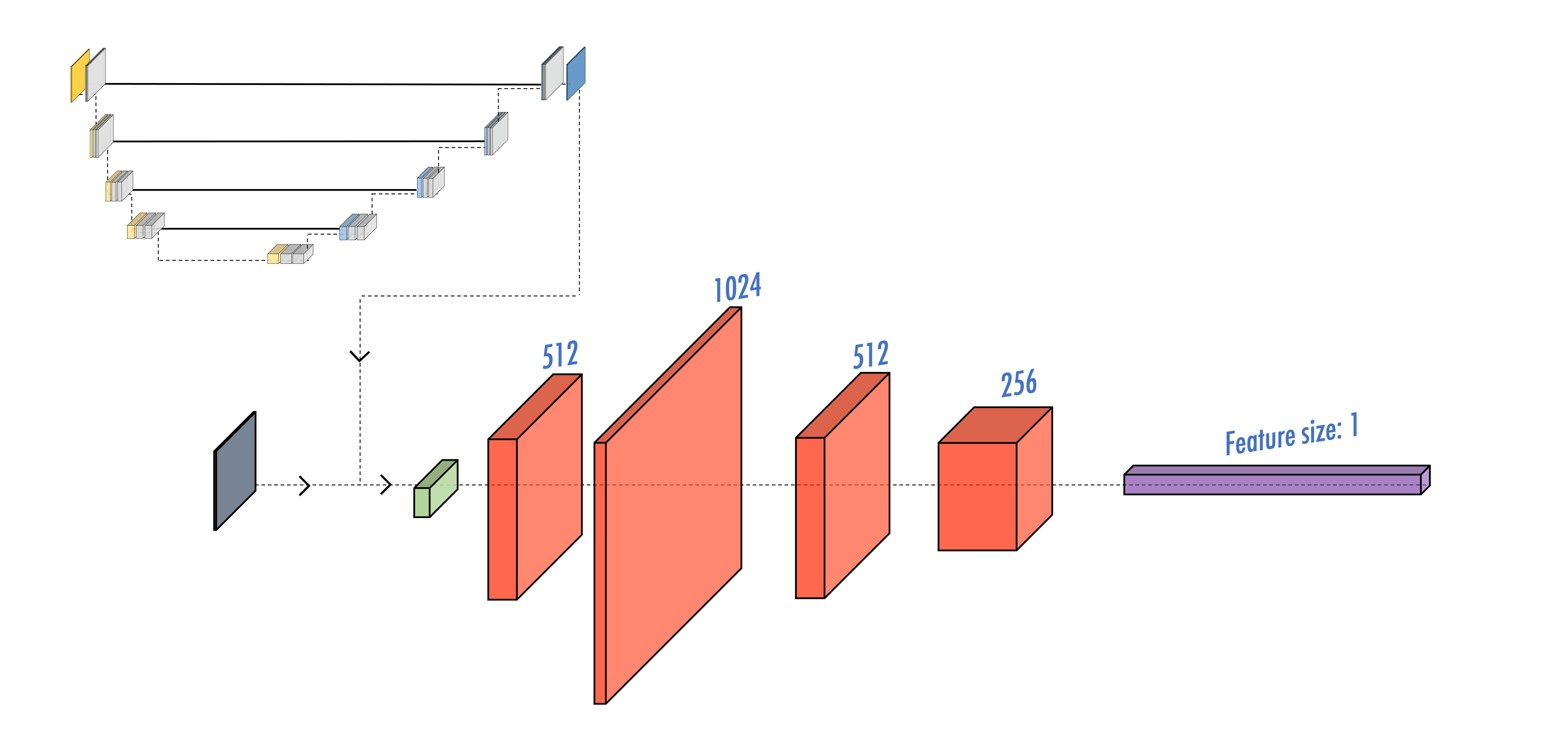}
         \caption{The D-4 GAN}
         \label{fig:gan1}
     \end{subfigure}
     \hfill
     \begin{subfigure}[b]{0.49\textwidth}
         \centering
            \includegraphics[width=\textwidth]{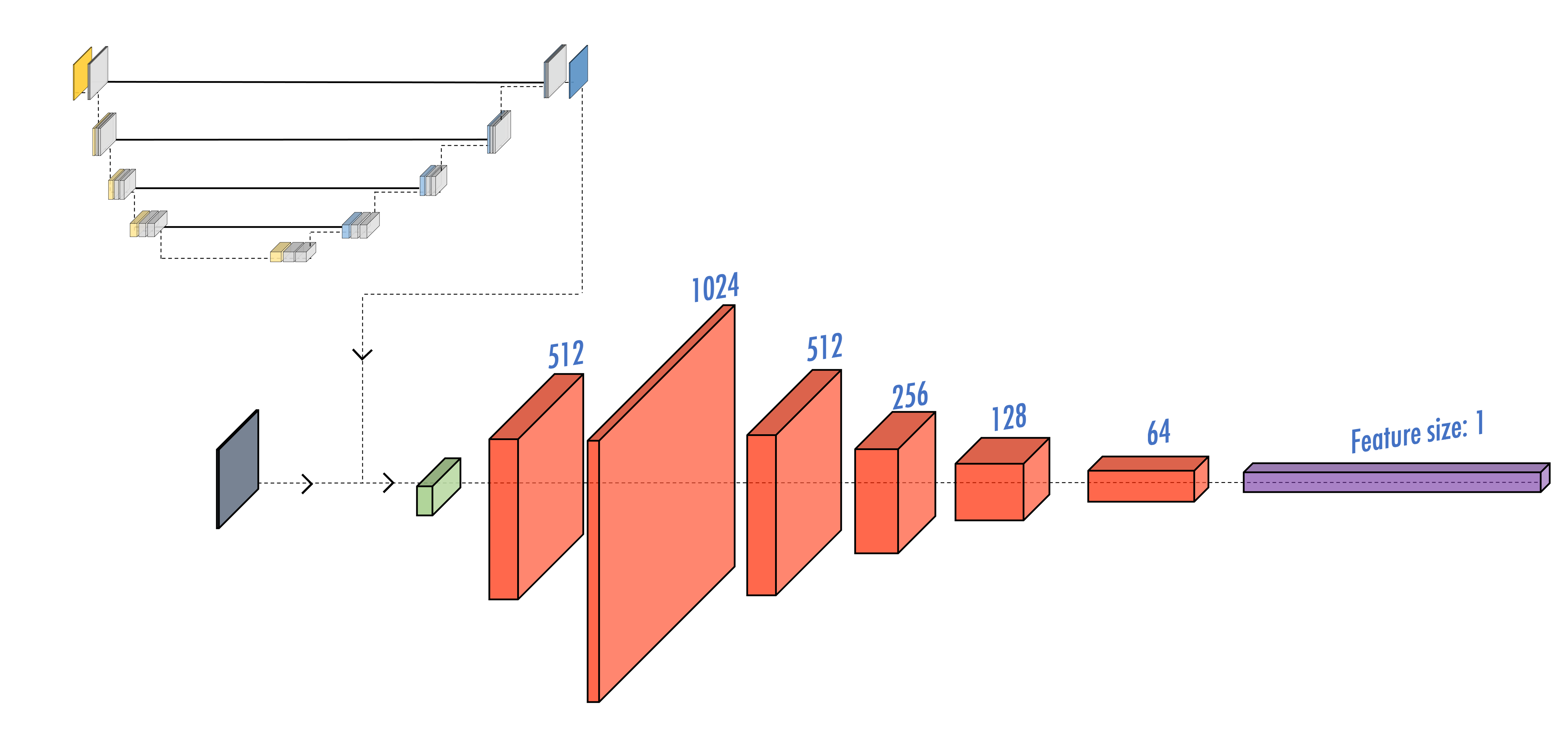}
         \caption{The D-6 GAN}
         \label{fig:gan2}
     \end{subfigure}
     \\
     \begin{subfigure}[b]{0.49\textwidth}
         \centering
            \includegraphics[width=\textwidth]{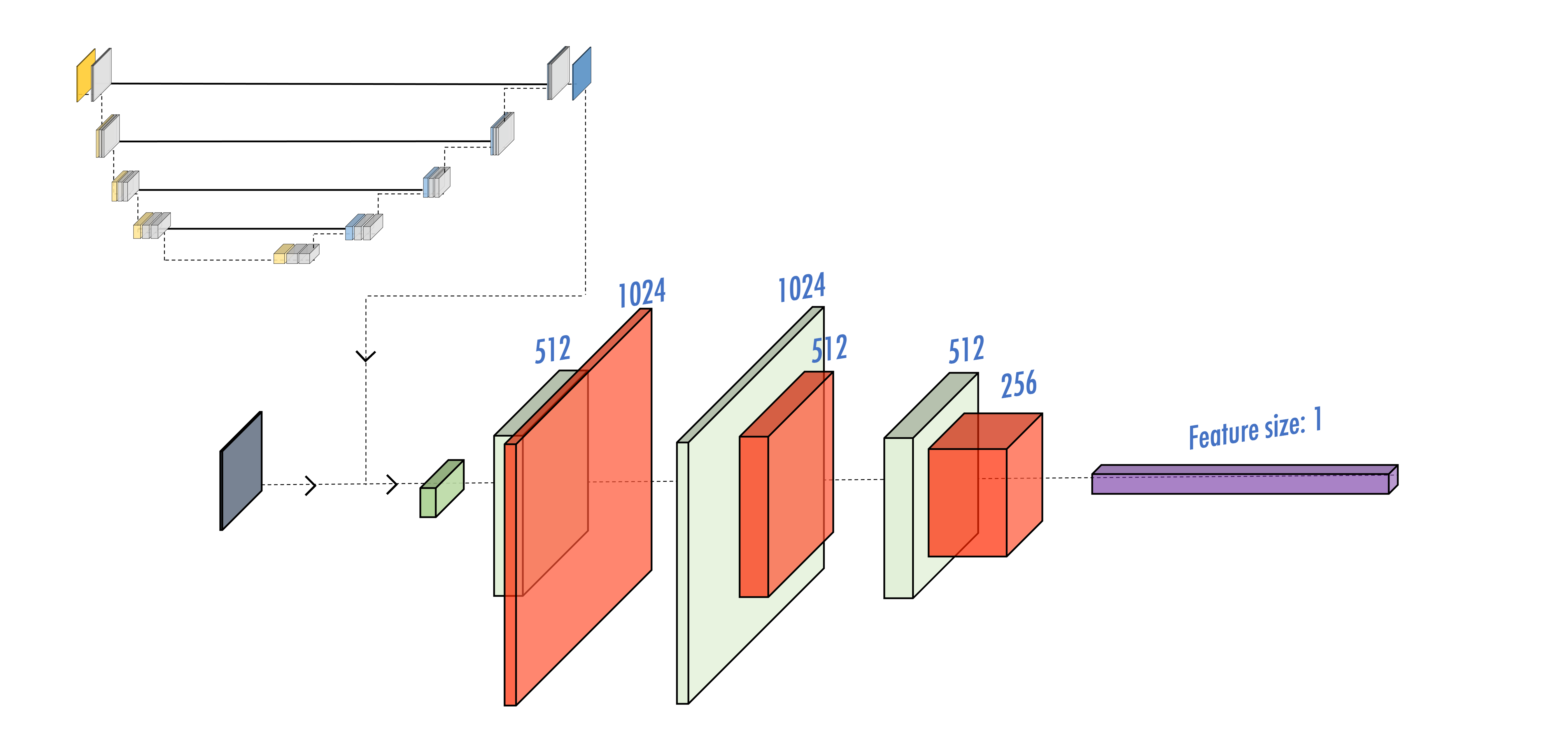}
         \caption{The D-4V GAN}
         \label{fig:gan3}
     \end{subfigure}
     \hfill
     \begin{subfigure}[b]{0.49\textwidth}
         \centering
            \includegraphics[width=\textwidth]{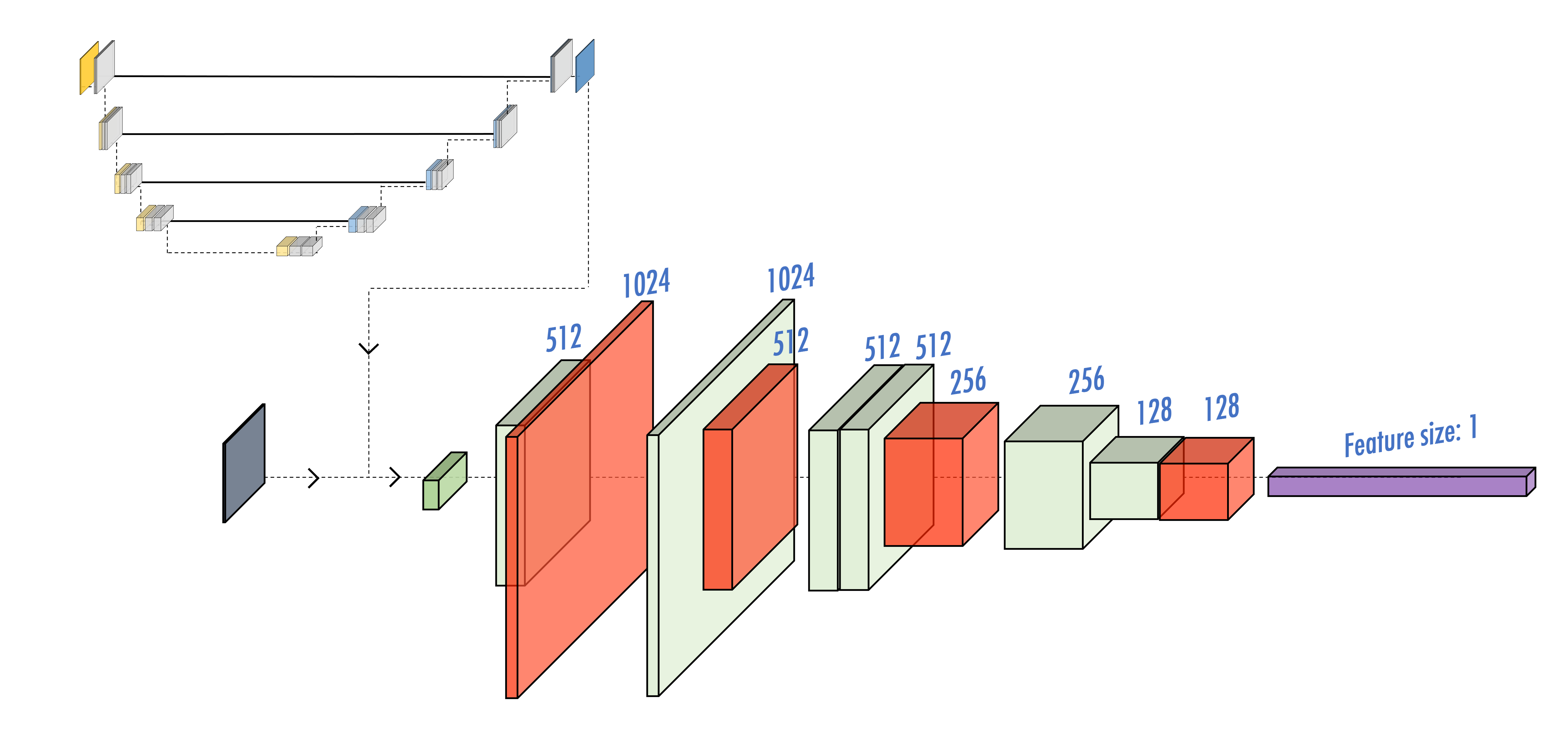}
         \caption{The D-5V GAN}
         \label{fig:gan4}
     \end{subfigure}
     \\
     \centering
     \begin{subfigure}[b]{\textwidth}
         \centering
            \includegraphics[width=\textwidth]{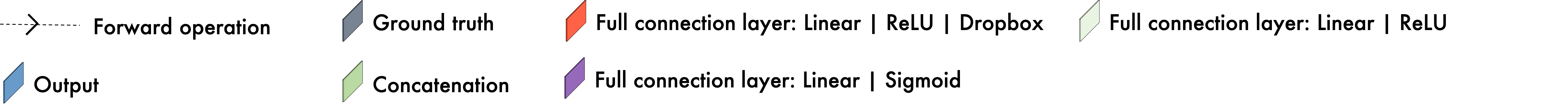}
     \end{subfigure}
    \caption{The proposed discriminators}
\end{figure*}

\subsection{The principle of GANs}

GANs are initially proposed as a generic architecture. For distinct objectives, e.g., classification, self-adaption, or semantic segmentation, the roles of generators and discriminators vary accordingly. As known, a GAN is composed of a generator and a discriminator (see Eq. \ref{eqGAN}); hence the training process should encompass both. First, the generator learns to generate plausible data. The instances that are created serve as negative training examples for the discriminator. Second, the discriminator learns to differentiate between the generator's actual and false data generated. The generator is penalized by the discriminator for providing improbable results. The generator learns to generate fake data by integrating discriminator feedback. It learns to manipulate the discriminator to identify its output as real. Generator training necessitates a closer integration of the generator and the discriminator than discriminator training necessitates. Random input, generator network, discriminator network, discriminator output, and generator loss are all items of the GAN that train the generator. In a GAN constructed for semantic segmentation, the generator can be interpreted as a network to produce the segmented images as the fake data; and the discriminator should evaluate the difference between the segmented images and the ground truths, i.e., the real data. The forward- and back-propagation are processed through both the generator and discriminator as a whole network, while only the generator, a partial GAN, addresses semantic segmentation. Therefore, it is possible to exclusively design a discriminator with a complex network structure for training, and the parameters in the generator can be optimized simultaneously. 

The U-net is adopted as the backbone of the generator in this research work. Approaches to enhancing the feature extraction of the U-net are explored. In such circumstances, the classifier is not requested to recognize massive sorts of objects but only the limited objects labeled in the ground truths. Hence, finding out a light but effective discriminator is the ultimate goal of this research. In addition, the implementation impedes this research work from creating a highly complicated discriminator as well. A GAN is a combination of two networks, which entails an enormous computing power in training and thus challenges the hardware configuration of a research institution. Due to the primary reasons, four discriminators are proposed for test and validation. These four discriminators are composed of fully connected layers to reduce the data processing load in training. The contingent impacts caused by different lengths of a discriminator and thickness of a single layer in discriminators are considered in the model establishment and will be investigated in the experiments. 

\subsection{The proposed discriminators}

Fig. \ref{fig:gan1} illustrates a D-4 GAN, in which the U-net serves as the generator, and it is combined with a discriminator with five fully connected layers (D-5). Each layer in the forepart encompasses three stages, a linear transformation to the incoming data, a ReLU activation function, and a dropout operation. After the image of ground truth and the segmented image are concatenated, the linear transformation is applied to extract features by scaling the data according to the given size requirements, such as the 512, 1024, 256, 128, 64, 1 in Fig. \ref{fig:gan1}. The transformed data should be activated by the ReLU function to maintenance the nonlinearity during data processing. The dropout operation is a practical technique to avoid the over-fitting problem. Given an empirical dropout rate, some elements of the input tensor are randomly zeroed with the probability of the dropout rate. The rear layer contains a sigmoid activation function and a linear transformer to switch the tensor with length 1. The D-6 GAN in Fig. \ref{fig:gan2} is named by its discriminator with seven layers. This discriminator is extended on the basis of the D-4 GAN and has an identical rear layer. In the forepart of the D-6 GAN, two fully connected layers are inserted with the linear transformation (256 to 128 and 128 to 64) for a more gentle descent of the tensor size.  

Inspired by the structure of the serial of VGG network series, the G-5 GAN is varied by duplication of linear and activation stages in the forepart of the discriminator, thus the variant is called the G-5V GAN. As Fig. \ref{fig:gan3} demonstrates, each layer is inserted by a linear transformation stage with same feature size of in- and out-put, and a ReLU activation function. The D-5V GAN (see Fig. \ref{fig:gan4}) is not a simple extension of the D-4V GAN. An extra complete fully connected layer is installed right before the rear part of the D-4V GAN. Moreover, the second stage in the second and the third layer in the D-5V GAN should be repetitively deducted.   

Researchers prefer to propose a novel architecture or a network and examine its better prediction results than other existing proposals. In this research, novel architectures and networks are proposed firstly for comparison to determine the best ones and applied for validation afterward. Thus, the design of the four proposed discriminators is oriented to the comparative study. The segmentation results of these four GANs can be cross-referenced to insight into the impacts from lengths and thicknesses.

\section{Implementation for the comparative study}\label{sec5}

The implementation of the proposed CNNs is executed through two steps. Initially, cell images are applied to compare the different combination of the U-net and attention mechanism in order to determine the CNNS owning the best performances. These selected networks are further employed for damage segmentation.

\subsection{Datasets}

The eventual objective of this paper is to determine the CNN models with the optimal performance in semantic segmentation and employ them to process the image of structural damage. As a matter of fact, images of structural damage (see Fig. \ref{fig:imp2}) usually involve more noise objects, e.g. walls, windows, backgrounds. Therefore, before the images of structural damage are fed into the CNN models, it makes sense to deduct a lightweight test, in which simple data can be applied to investigate the preceding models initially with less time consumption, however, the prediction quality still can be well examined. Cell images (see Fig. \ref{fig:imp11} and \ref{fig:imp12}) with similar graphical features like facade damage but less interference in images fulfill the data requirements of a lightweight test since these graphic data contain simple interference noise such as organelles but distinct cytoderm for segmentation. The lightweight test is essentially a comparative study to investigate the performances respectively of attention mechanism and the discriminator in cell image segmentation. 

In the lightweight test, 24 images are adopted as the input images, the corresponding 24 images of ground truth are labels. The other five images are utilized as the text data; the prediction results of segmentation will be compared with these five images of ground truth via PA and mIoU. The four architectures, i.e., the U-net, the attention U-net, the advanced attention U-net, and the full attention U-net (see Fig. \ref{fig:attn1}, \ref{fig:attn2}, \ref{fig:attn3}, and \ref{fig:attn4}) are trained with the same configuration of hyperparameters. Such configuration does not incur these models to reach their own highest performance. However, on the basis of the unified standard of training hyperparameters, the segmentation results are sufficiently plausible to reflect the strengthens and weaknesses distinctly.  

The dataset for facade damage segmentation applied the images of physics-based graphics models. These models are generated in a virtual digital environment and annotated with spalling areas, cracks, and exposed reinforcement bars to present the damage in facades after an earthquake or long-term weathering. The procedure of image acquisition also imitates the view angle and the movement trajectory of UAVs. This implementation focuses on the semantic segmentation of spall in the building facade. Therefore, the dataset is composed a vast number of images and the corresponding labels of spalling areas (see Fig. \ref{fig:imp2}). The large amount of the original images have a resolution of 1980$\times$1080. For an efficient implementation, 1,000 representative images and the corresponding ground truths are selected and resized into 512$\times$512.



\subsection{Training configuration}

Since the configuration for training is not a research issue in this paper, it is hence empirical to select a loss function, an optimization algorithm, and metrics. Hereby, the RMSprop algorithm is chosen for the optimizer, the Binary Cross Entropy (BCE) with logits loss works as the loss function. Pixel Accuracy and the mean Intersection over Union (mIoU) are employed to evaluated the results of semantic segmentation. 

\subsubsection{Optimization algorithm}

Gradient descent is one of the most prominent optimization methods, and it is by far the most frequent method for optimizing neural networks. Nowadays, every available Deep Learning framework (e.g., Tansorflow \citep{tensorflow2015-whitepaper}, Pytorch \citep{NEURIPS2019_9015}, Keras \citep{chollet2015keras}) includes numerous methods for optimizing gradient descent. However, since practical descriptions of their benefits and limitations are difficult to come by, these algorithms are frequently employed as black-box optimizers. Gradient descent is a method for minimizing an objective function parameterized by model parameters by updating the parameters in the opposite direction as the gradient of the objective function. The learning rate determines the magnitude of the steps taken to attain a minimum.

RMSprop in Eq. \ref{eqOp1} to Eq. \ref{eqOp4} is a gradient-based optimization method used in neural network training. Gradients of extremely complicated functions, such as neural networks, have a propensity to disappear or explode as data passes through the function. Rmsprop is a stochastic approach for mini-batch learning. RMSprop addresses the aforementioned issue by normalizing the gradient using a moving average of squared gradients. This normalization balances the step size, reducing the step size for big gradients to prevent bursting and increasing the step size for small gradients to avoid disappearing. 

\begin{equation} \label{eqOp1}
\begin{split}
\nu _{t} = \beta _{1}\cdot \nu _{t-1}-\left ( 1-\beta _{1} \right )\cdot g_{t}
\end{split}
\end{equation}
\begin{equation} \label{eqOp2}
\begin{split}
s _{t} = \beta _{2}\cdot s _{t-1}-\left ( 1-\beta _{2} \right )\cdot g_{t}^{2}
\end{split}
\end{equation}
\begin{equation} \label{eqOp3}
\begin{split}
\Delta \omega _{t}=-\eta \frac{\nu _{t}}{\sqrt{s_{t}+\epsilon }}\cdot g_{t}
\end{split}
\end{equation}
\begin{equation} \label{eqOp4}
\begin{split}
\omega _{t+1}= \omega _{t}+\Delta \omega _{t}
\end{split}
\end{equation}

\begin{equation*}
\begin{split}
\eta :&\ initial\ learning\ rate \\
g_{t}:&\ gradient\ at\ time\ t\ along\ \omega _{j} \\
\nu _{t}:&\ exponential\ average\ of\ gradients\ along\ \omega _{j} \\ 
s _{t}:&\ exponential\ average\ of\ squares\ of\ gradients \\
 &\ along\ \omega _{j} \\ 
\beta _{1},\beta _{2}:&\ hyperparameters \\
\end{split}
\end{equation*}

\subsubsection{Loss function}

During training a CNN, the degree of deviation between the predicted and input data is represented by loss calculated by loss functions. The calculated loss affects the backpropagation over training epochs. A larger loss indicates the more significant difference between the inputs and predictions. The loss should be minimized epoch by epoch in training iteration as much as possible. Thus a minor loss is always preferred in model training. The BCE with logits loss is adopted in this research work. This loss represented in Eq. \ref{eqLossCNN} combines a sigmoid layer and the BCE loss \citep{torch_contributors_bceloss_2019} in one single class.

\begin{equation} \label{eqLossCNN}
\begin{split}
l_{n} = -\omega _{n}\left \{ t_{n} \cdot log\ \sigma \left (x_{n}  \right ) + \left ( 1-t_{n}\right )\cdot log\ \left [ 1-\sigma \left ( x_{n} \right ) \right ] \right \}
\end{split}
\end{equation}

\begin{equation*}
\begin{split}
l_{n}:&\ caculatd\ loss\ in\ the\ n^{th}\ loop \\
\omega _{n}:&\ weight\ parameter in\ the\ n^{th}\ loop \\
x_{n}:&\ precsion of correct prediction in\ the\ n^{th}\ loop \\
t_{n}:&\ target\ in\ the\ n^{th}\ loop \\
\sigma \left (  \right ):&\ sigmoid\ acvtivation\ function\ in\ the\ n^{th}\ loop \\
\end{split}
\end{equation*}

\subsubsection{Metrics}

PA in Eq. \ref{eqMetr1} provides the percentage of pixels in the images identified adequately as a metric for evaluating semantic segmentation. This metric is usually provided independently for each class as well as globally across all classes. In the assessment of per-class pixel accuracy, a binary mask is evaluated. A true positive indicates a properly projected pixel to belong to the given class according to the target mask. In contrast, a true negative represents a correctly detected as not belonging to the given class. However, when the class representation is minimal inside the image, this metric can often produce deceptive results since the measure is skewed in reporting how effectively you detect negative cases.

\begin{equation} \label{eqMetr1}
\begin{split}
PA = \frac{TP + TN}{TP + TN + FP + FN}
\end{split}
\end{equation}

\begin{equation*}
\begin{split}
TP:&\ true\ positive \\
TN:&\ true\ negative \\
FP:&\ false\ positive \\
FP:&\ false\ negative \\
\end{split}
\end{equation*}

The IoU metric in Eq. \ref{eqMetr2} is a way for quantifying the percentage overlap between the target mask and our prediction output. The IoU metric is calculated by dividing the number of pixels shared by the target and prediction masks by the total number of pixels contained in both masks. 

\begin{equation} \label{eqMetr2}
\begin{split}
IoU = \frac{target \cap prediction}{target \cup prediction }
\end{split}
\end{equation}

The intersection \(target \cap predictio \) is made up of pixels from both the prediction and ground truth masks, whereas the union \(target \cup prediction \) is made up of pixels from either the prediction or target mask. The IoU score is generated individually for each class and averaged across all classes to produce the semantic segmentation prediction's global, namely mean IoU (mIoU) score.

\subsection{A lightweight test using cell images}

\begin{figure*}[bh]
     \begin{subfigure}[b]{0.16\textwidth}
         \centering
            \includegraphics[width=\textwidth]{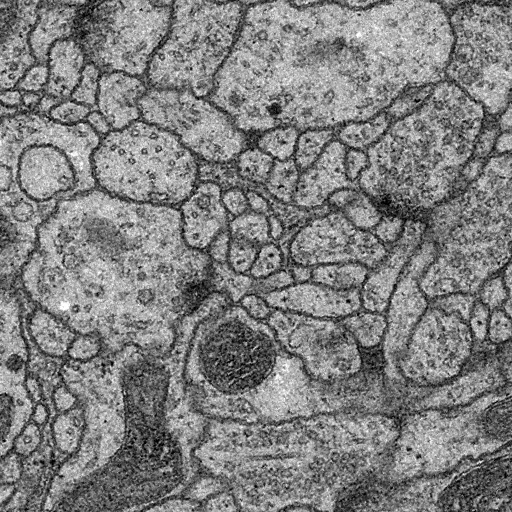}
         \caption{Ground truth}
         \label{fig:cvlwts12}
     \end{subfigure}
     \hfill
     \begin{subfigure}[b]{0.16\textwidth}
         \centering
            \includegraphics[width=\textwidth]{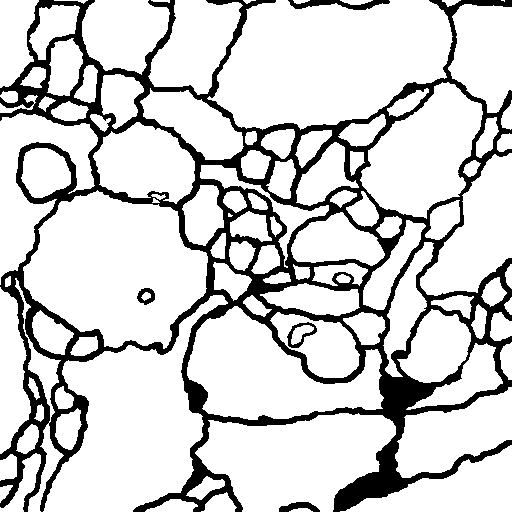}
         \caption{Input image}
         \label{fig:cvlwts11}
     \end{subfigure}
     \hfill
     \begin{subfigure}[b]{0.16\textwidth}
         \centering
            \includegraphics[width=\textwidth]{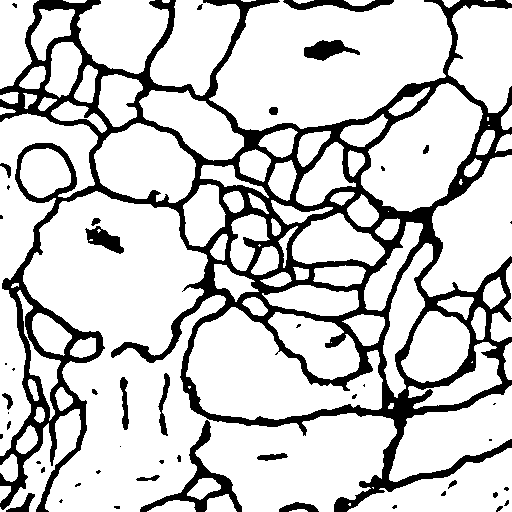}
         \caption{U-net}
         \label{fig:cvlwts13}
     \end{subfigure}
     \hfill
     \begin{subfigure}[b]{0.16\textwidth}
         \centering
            \includegraphics[width=\textwidth]{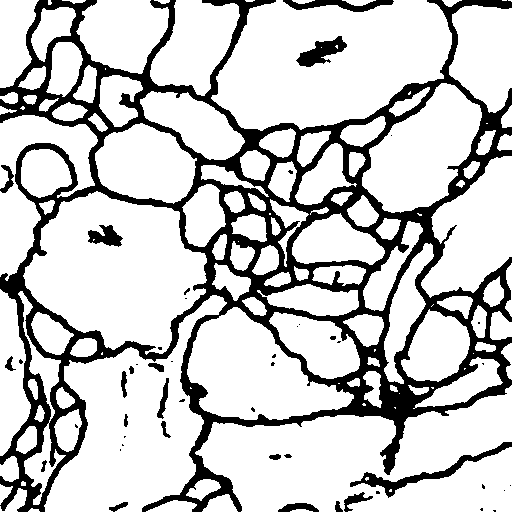}
         \caption{Attn.U-net}
         \label{fig:cvlwts14}
     \end{subfigure}
     \hfill
     \begin{subfigure}[b]{0.16\textwidth}
         \centering
            \includegraphics[width=\textwidth]{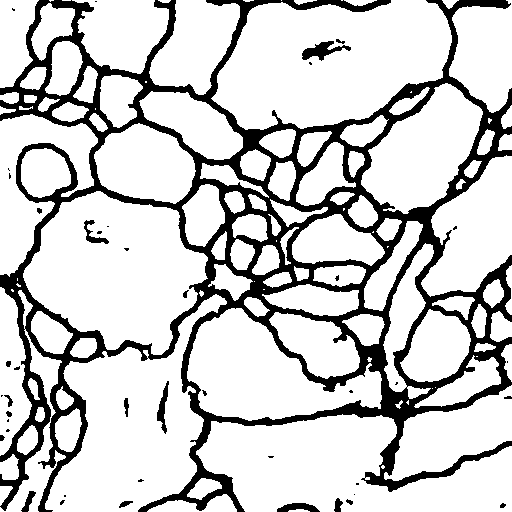}
         \caption{Adv. attn. U-net}
         \label{fig:cvlwts15}
     \end{subfigure}
     \hfill
     \begin{subfigure}[b]{0.16\textwidth}
         \centering
            \includegraphics[width=\textwidth]{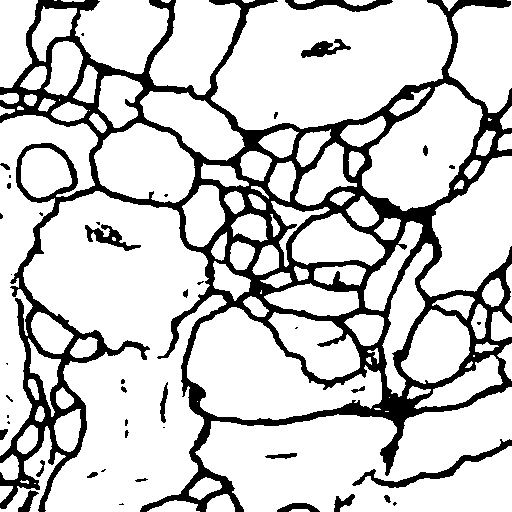}
            \caption{Full attn. U-net}
            \label{fig:cvlwts16}
     \end{subfigure}
    \caption{Data and result examples for the lightweight test for attention mechanism}
    \label{fig:imp11}
\end{figure*}

\begin{figure*}[hb]
     \begin{subfigure}[b]{0.16\textwidth}
         \centering
            \includegraphics[width=\textwidth]{24.png}
         \caption{Ground truth}
     \end{subfigure}
     \hfill
     \begin{subfigure}[b]{0.16\textwidth}
         \centering
            \includegraphics[width=\textwidth]{24_1.png}
         \caption{Input image}
     \end{subfigure}
     \hfill
     \begin{subfigure}[b]{0.16\textwidth}
         \centering
            \includegraphics[width=\textwidth]{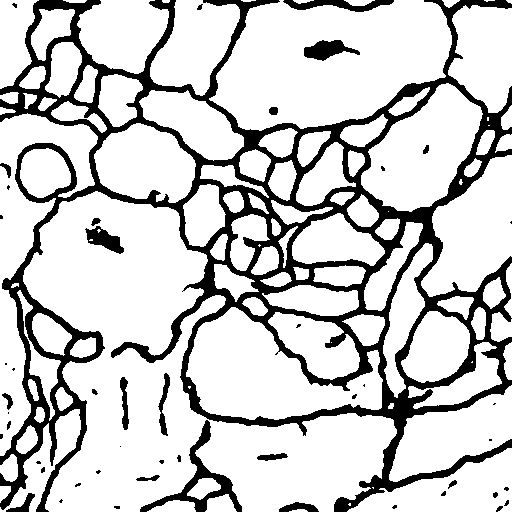}
         \caption{D-4 GAN}
     \end{subfigure}
     \hfill
     \begin{subfigure}[b]{0.16\textwidth}
         \centering
            \includegraphics[width=\textwidth]{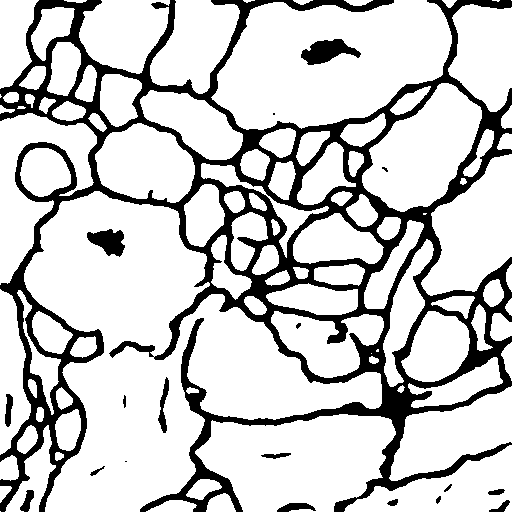}
         \caption{D-6 GAN}
     \end{subfigure}
     \hfill
     \begin{subfigure}[b]{0.16\textwidth}
         \centering
            \includegraphics[width=\textwidth]{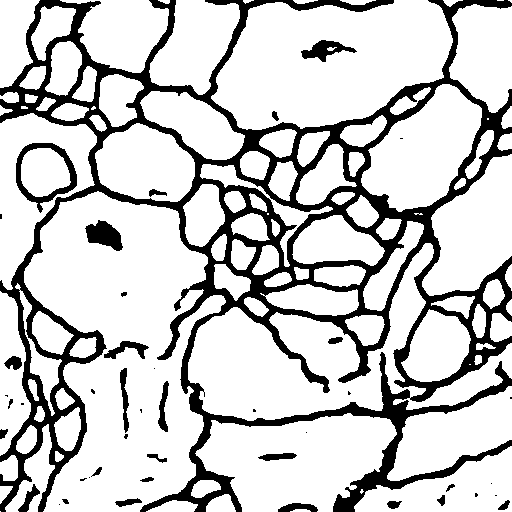}
         \caption{D-4V GAN}
     \end{subfigure}
     \hfill
     \begin{subfigure}[b]{0.16\textwidth}
         \centering
            \includegraphics[width=\textwidth]{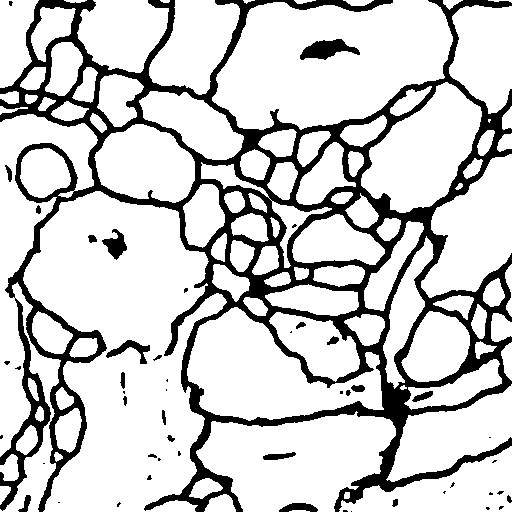}
            \caption{D-5V GAN}
     \end{subfigure}
    \caption{Data and result examples for the lightweight test for discriminators}
    \label{fig:imp12}
\end{figure*}

Tab. \ref{tab:cvtest1} displays the result evaluation of the lightweight test. PA emphasizes the correct location of the pixels in the prediction. According to the metric, the full attention U-net and advanced attention exhibit approximately 0.3\% to 0.4\% higher than the U-net and the attention U-net. The mIoU is a more comprehensive evaluation considering the locations of right and wrong pixels in the prediction results. Though the prediction enhancement caused by the attention mechanism is distinctly presented, it is uncertain that the prediction quality is positively correlated with the distribution density of the attention mechanism. The mIoU of advanced U-net, 78.22\%, shows a significant advantage, as the full attention U-net with more attention gate and more complex combination of skip connections has merely the mIoU of 77.97\%. The examples in Fig. \ref{fig:imp11} endorse the data in Tab. \ref{tab:cvtest1} as well. Fewer and lighter noise spots appear in Fig. \ref{fig:cvlwts15} and \ref{fig:cvlwts16} than those in Fig. \ref{fig:cvlwts13} and \ref{fig:cvlwts14}. The cross-comparison of Fig. \ref{fig:cvlwts15} and \ref{fig:cvlwts16} indicates that the advanced attention U-net is able to predict the cytoderm with higher completeness. Therefore, consequently, the advanced attention U-net performs best due to its appropriate distribution of attention gates and utilization of skip connections.

\begin{table}[htbp]
\caption{Result evaluation of the lightweight test}
\label{tab:cvtest1}
\centering
    \begin{threeparttable}
        \begin{tabular}{lcccccc}
            \toprule
             & \textbf{PA\tnote{$^{\rm a}$}} (\%) & \textbf{$\Delta$}\tnote{$^{\rm b}$} (\%) & \textbf{mIoU}\tnote{$^{\rm c}$} (\%) & \textbf{$\Delta$} (\%)  \\
             \midrule
            U-net & 91.55 & - & 77.56 & -  \\
            Attn. U-net \tnote{$^{\rm d}$} & 91.50 & -0.055 & 77.73 & +0.22 \\
            Adv. attn. U-net \tnote{$^{\rm e}$} & \textbf{91.88} & \textbf{+0.36} & \textbf{78.22} & \textbf{+0.85} \\
            Full attn. U-net \tnote{$^{\rm f}$} & 91.85 & +0.33 & 77.97 & +0.53 \\
            \bottomrule
        \end{tabular}
        \quad
        \begin{tabular}{lcccccc}
            \toprule
             & \textbf{PA} (\%) & \textbf{$\Delta$} (\%) & \textbf{mIoU}\ (\%) & \textbf{$\Delta$} (\%)  \\
             \midrule
            GAN + D-4 & 91.69 & - & 78.00 & -  \\
            GAN + D-6 & \textbf{91.79} & \textbf{+0.11} & \textbf{78.23} & \textbf{+0.29} \\
            GAN + D-4V & 91.44 & -0.27 & 77.63 & -0.47 \\
            GAN + D-5V & 91.74 & +0.05 & 77.72 & -0.35 \\
            \bottomrule
        \end{tabular}
    
        \begin{tablenotes}
        \item[$^{\rm a}$] PA: Pixel Accuracy
        \item[$^{\rm b}$] $\Delta$: the increment compared with the results from the U-net
        \item[$^{\rm c}$] mIoU: the mean Intersection over Union
        \item[$^{\rm d}$] Attn. U-net: the attention U-net
        \item[$^{\rm e}$] Adv. attn. U-net: the advanced attention U-net
        \item[$^{\rm e}$] Full attn. U-net: the full attention U-net
        \end{tablenotes}
    \end{threeparttable}
\end{table}

The attached discriminator in a GAN can alert the loss function's tendency to convergence. The four proposed discriminators are coupled with the U-net for training and testing. The discriminator comparative study investigates the four discriminators constructed with different layer numbers and thicknesses. The two metrics of the D-6 GAN are 0.11\% and 0.29\% (see Tab. \ref{tab:cvtest1}) higher than those two values of the D-4 GAN. Moreover, the results of the D-5V GAN are relatively better than the D-4V GAN, even though the layers in the two architectures are not completely conceived. In the condition that two discriminators are established with the same layer thickness, the discriminator with more layers segments the images with higher accuracy in PA and mIoU. Two discriminators (e.g., the D-4 GAN and the D-4V GAN) with the same length can deliver different results for the different layer designs. Thus, the more operations are inserted in a discriminator's layer, the worse metrics of the discriminator can be obtained. The illustrations in Fig. \ref{fig:imp12} reflect the calculated metrics. The predictions of the D-6 and D-5V GANs include fewer interference objects in cells and more details of cell walls.     
According to the lightweight test, the strategy of advanced attention materializes a significant promotion in semantic segmentation; the generative adversarial network with discriminator including six fully connected layers wins the competition with the other three proposed ones. Therefore, these two architectures (see Fig. \ref{fig:attn3} and \ref{fig:gan2}) are employed in the experiment. Besides, a complex U-net, which is composed of the U-net, the advanced attention mechanism, and the discriminator with six layers, is constructed and experimented with the images of structural damage for validation to explore the combined effects in semantic segmentation. The U-net (see Fig. \ref{fig:attn1}) participates in the experiment, too, as the reference for result analysis.

\subsection{Damage segmentation for on-site inspection}

\begin{figure*}[hbtp]
     \centering
        \includegraphics[width=0.7\textwidth]{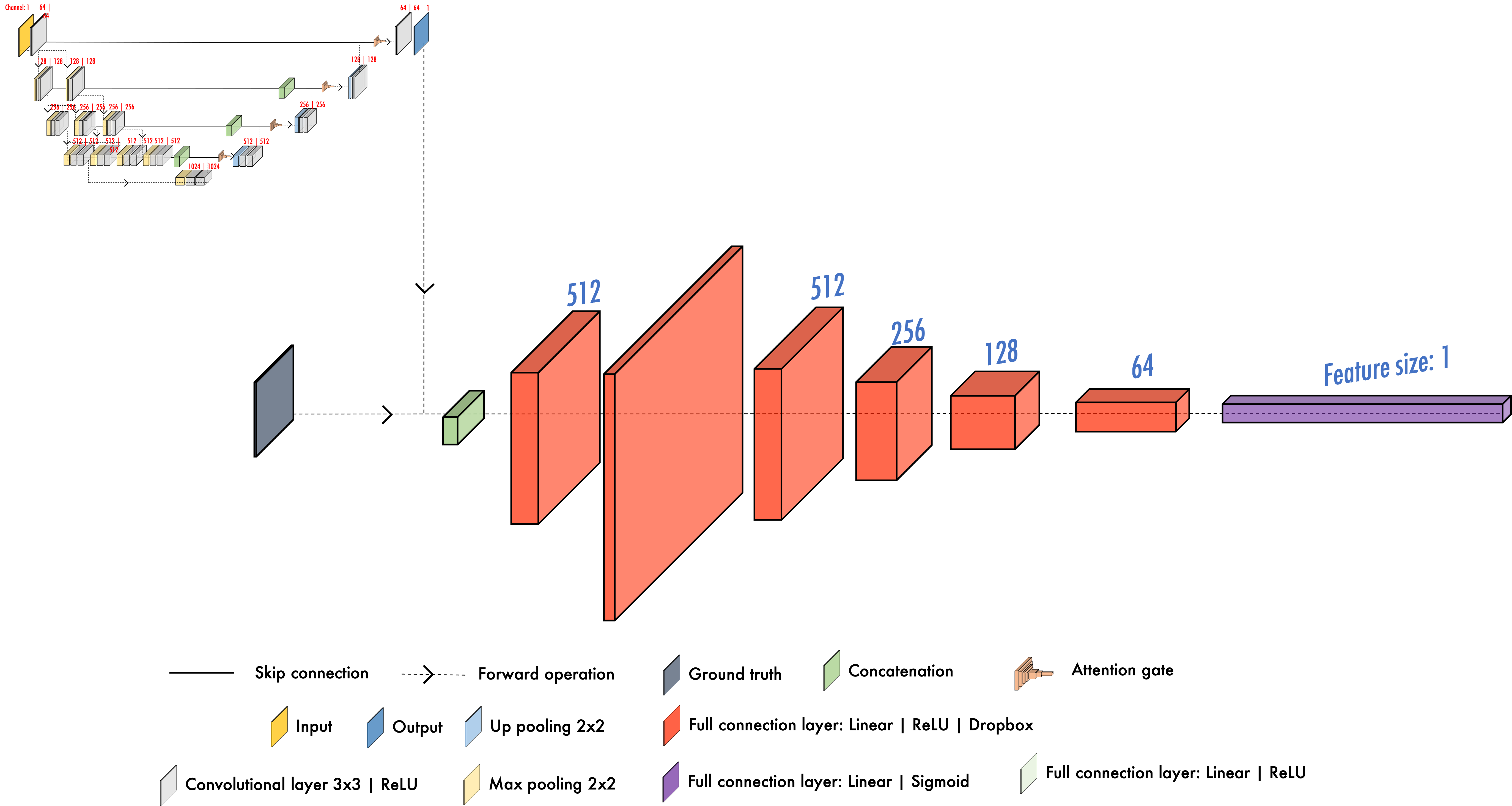}
     \caption{Com. network 1}
     \label{fig:ComNet1}
\end{figure*}

\begin{figure*}[ht]
     \centering
     \begin{subfigure}[b]{0.16\textwidth}
         \centering
            \includegraphics[width=\textwidth]{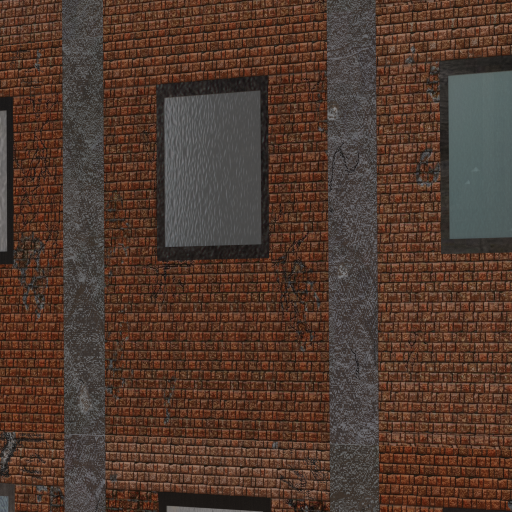}
     \end{subfigure}
     \hfill
     \begin{subfigure}[b]{0.16\textwidth}
         \centering
            \includegraphics[width=\textwidth]{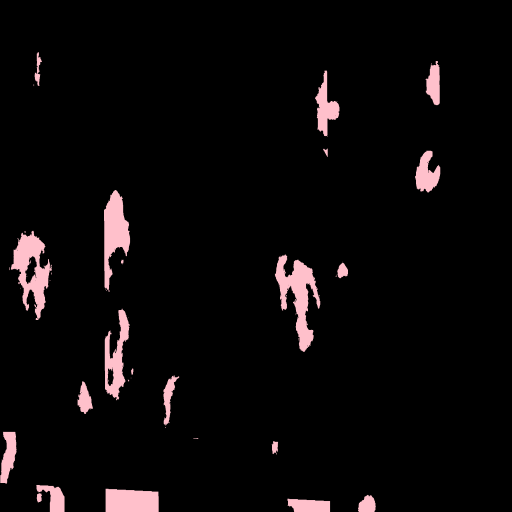}
     \end{subfigure}
     \hfill
     \begin{subfigure}[b]{0.16\textwidth}
         \centering
            \includegraphics[width=\textwidth]{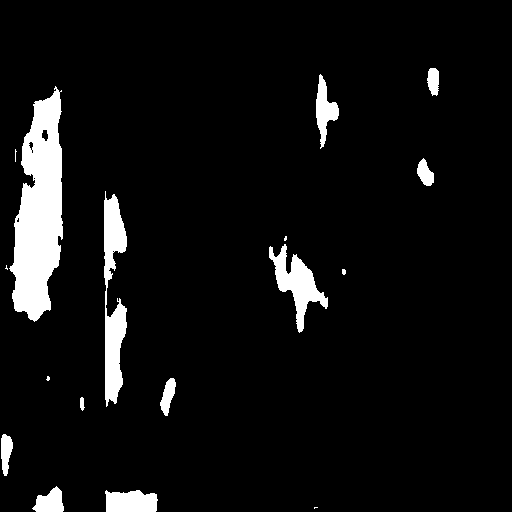}
     \end{subfigure}
     \hfill
     \begin{subfigure}[b]{0.16\textwidth}
         \centering
            \includegraphics[width=\textwidth]{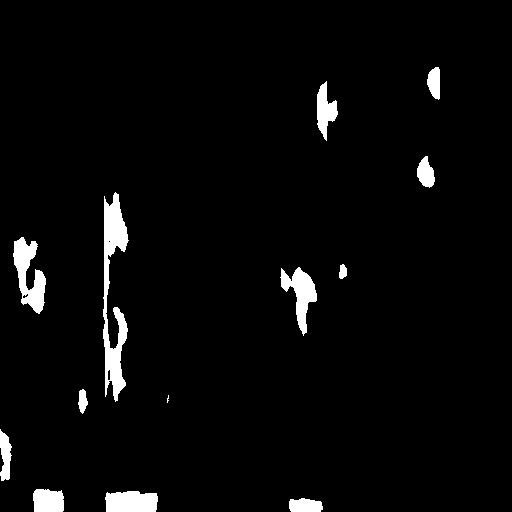}
     \end{subfigure}
     \hfill
     \begin{subfigure}[b]{0.16\textwidth}
         \centering
            \includegraphics[width=\textwidth]{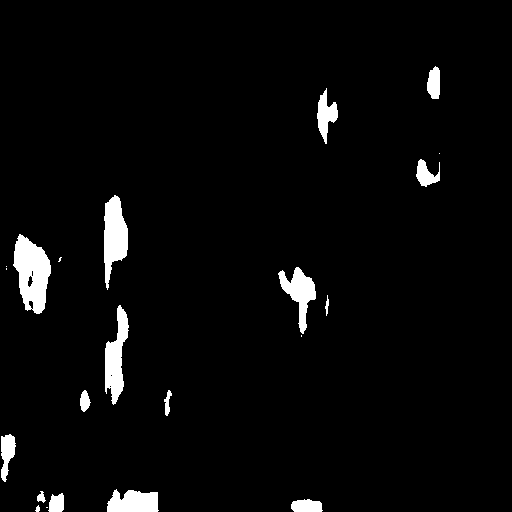}
     \end{subfigure}
     \hfill
     \begin{subfigure}[b]{0.16\textwidth}
         \centering
            \includegraphics[width=\textwidth]{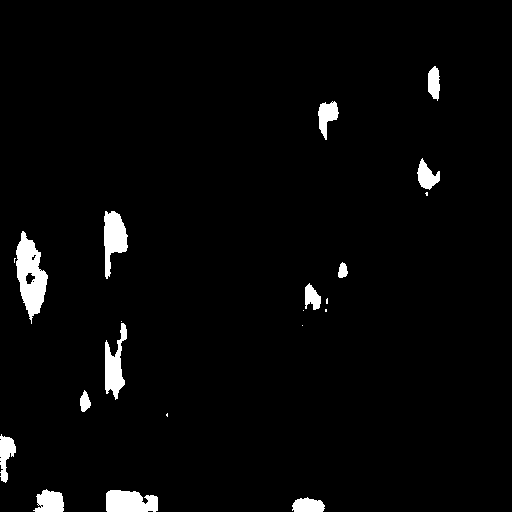}
     \end{subfigure}
    \\
     \begin{subfigure}[b]{\textwidth}
     \end{subfigure}
    \\
     \begin{subfigure}[b]{0.16\textwidth}
         \centering
            \includegraphics[width=\textwidth]{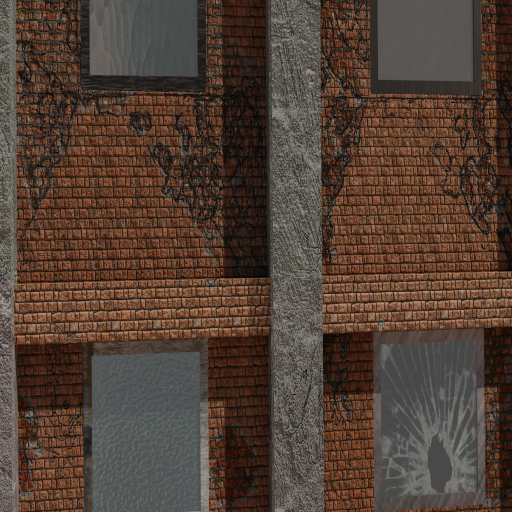}
     \end{subfigure}
     \hfill
     \begin{subfigure}[b]{0.16\textwidth}
         \centering
            \includegraphics[width=\textwidth]{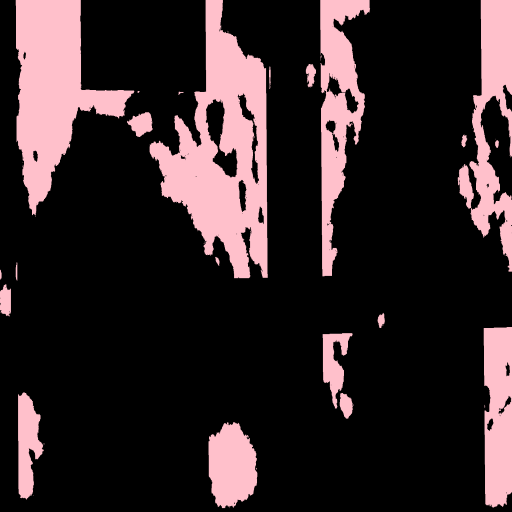}
     \end{subfigure}
     \hfill
     \begin{subfigure}[b]{0.16\textwidth}
         \centering
            \includegraphics[width=\textwidth]{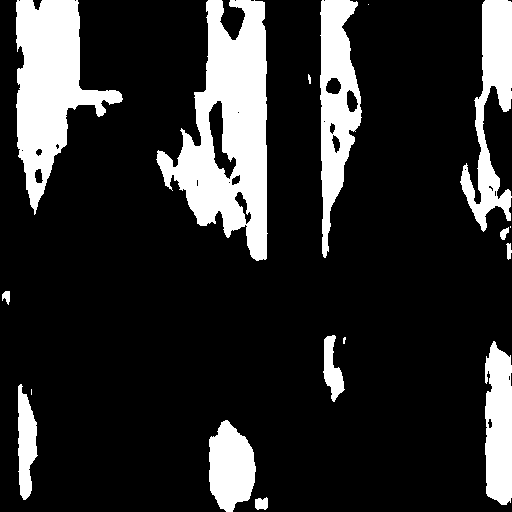}
     \end{subfigure}
     \hfill
     \begin{subfigure}[b]{0.16\textwidth}
         \centering
            \includegraphics[width=\textwidth]{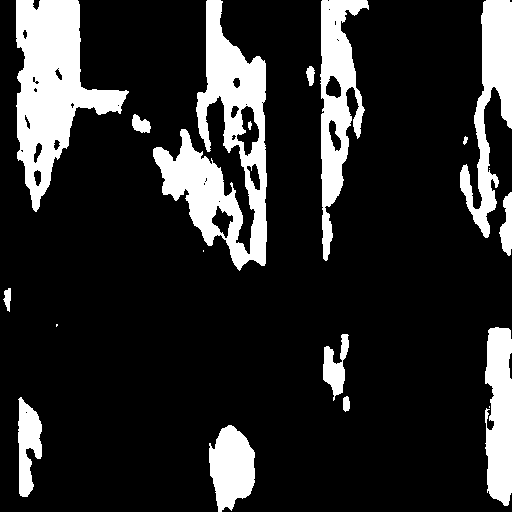}
     \end{subfigure}
     \hfill
     \begin{subfigure}[b]{0.16\textwidth}
         \centering
            \includegraphics[width=\textwidth]{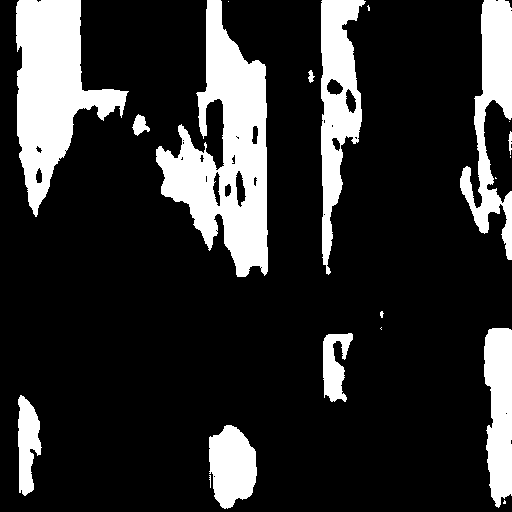}
     \end{subfigure}
     \hfill
     \begin{subfigure}[b]{0.16\textwidth}
         \centering
            \includegraphics[width=\textwidth]{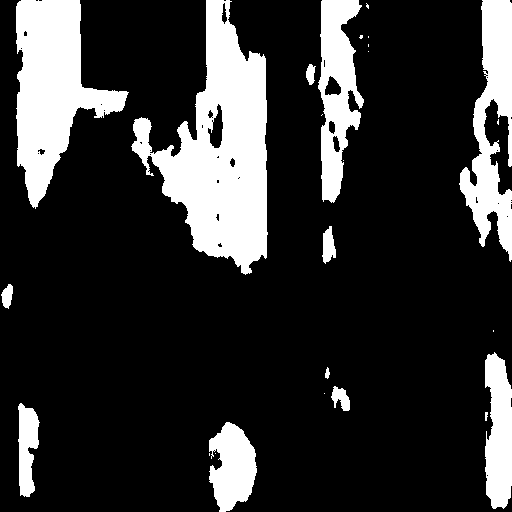}
     \end{subfigure}
     \\
     \begin{subfigure}[b]{\textwidth}
     \end{subfigure}
     \\
     \begin{subfigure}[b]{0.16\textwidth}
         \centering
            \includegraphics[width=\textwidth]{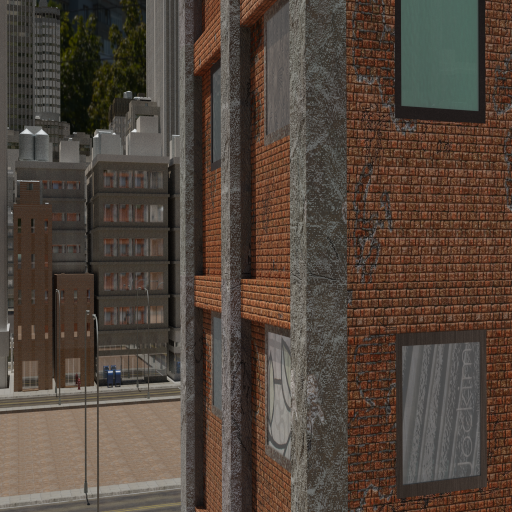}
     \end{subfigure}
     \hfill
     \begin{subfigure}[b]{0.16\textwidth}
         \centering
            \includegraphics[width=\textwidth]{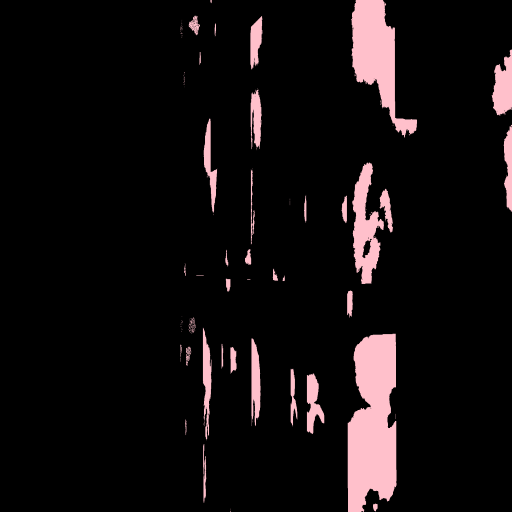}
     \end{subfigure}
     \hfill
     \begin{subfigure}[b]{0.16\textwidth}
         \centering
            \includegraphics[width=\textwidth]{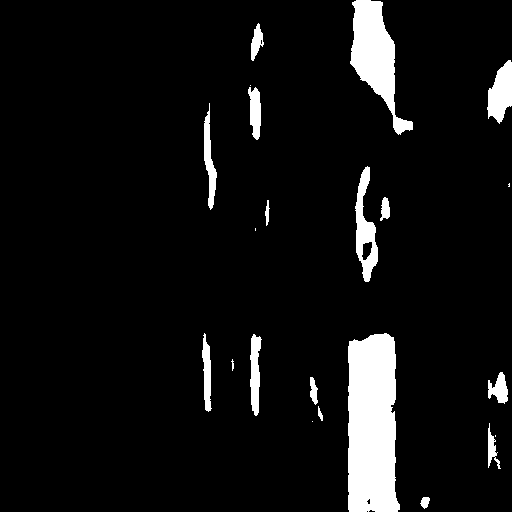}
     \end{subfigure}
     \hfill
     \begin{subfigure}[b]{0.16\textwidth}
         \centering
            \includegraphics[width=\textwidth]{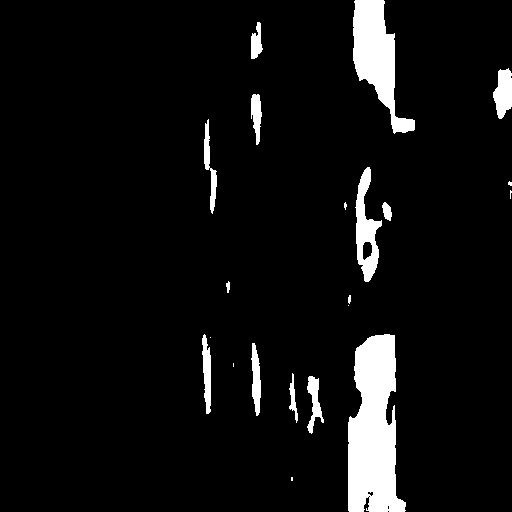}
     \end{subfigure}
     \hfill
     \begin{subfigure}[b]{0.16\textwidth}
         \centering
            \includegraphics[width=\textwidth]{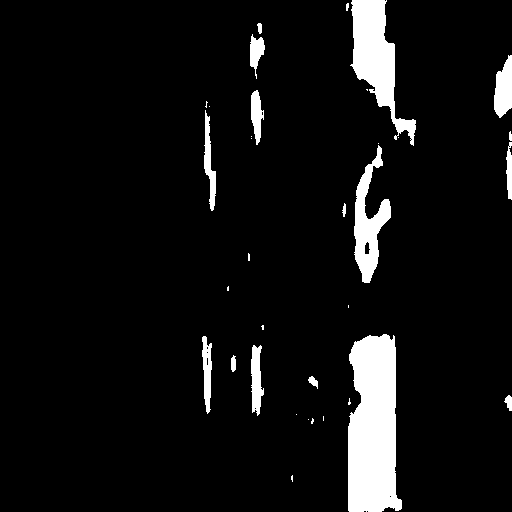}
     \end{subfigure}
     \hfill
     \begin{subfigure}[b]{0.16\textwidth}
         \centering
            \includegraphics[width=\textwidth]{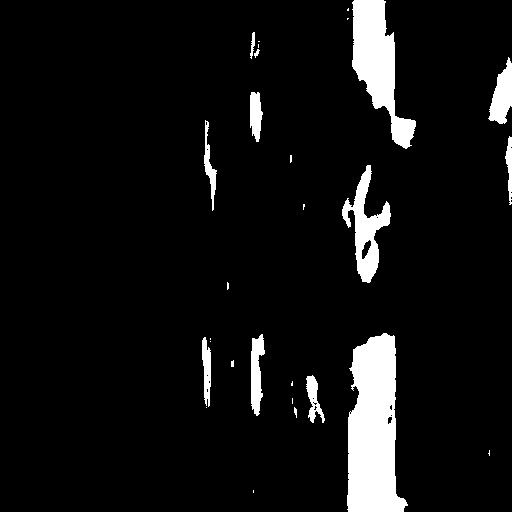}
     \end{subfigure}
     \\
     \begin{subfigure}[b]{\textwidth}
     \end{subfigure}
     \\
     \begin{subfigure}[b]{0.16\textwidth}
         \centering
            \includegraphics[width=\textwidth]{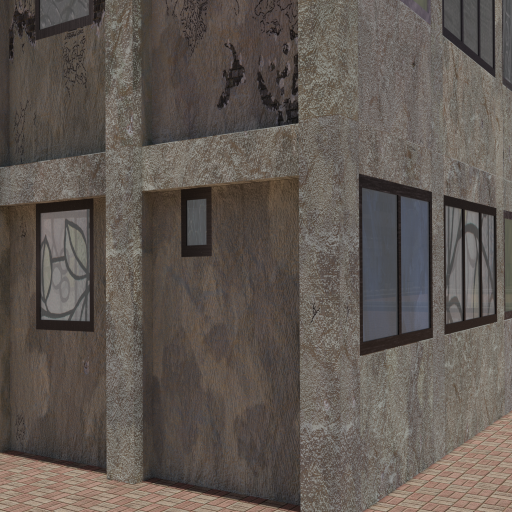}
     \end{subfigure}
     \hfill
     \begin{subfigure}[b]{0.16\textwidth}
         \centering
            \includegraphics[width=\textwidth]{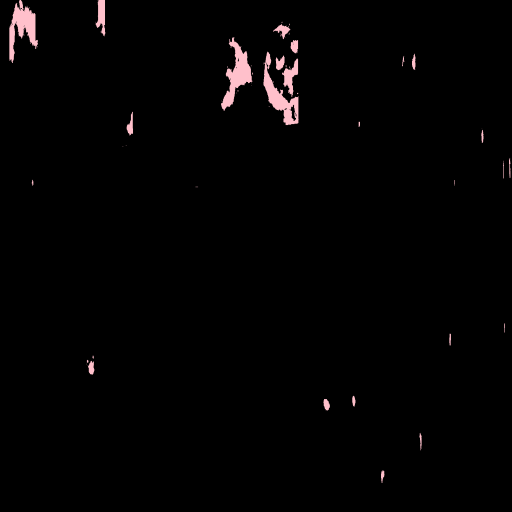}
     \end{subfigure}
     \hfill
     \begin{subfigure}[b]{0.16\textwidth}
         \centering
            \includegraphics[width=\textwidth]{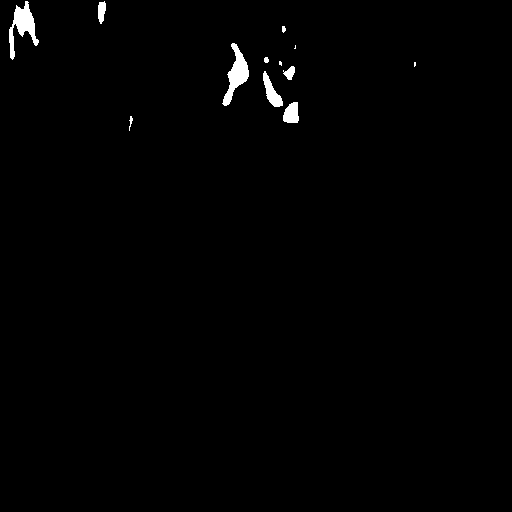}
     \end{subfigure}
     \hfill
     \begin{subfigure}[b]{0.16\textwidth}
         \centering
            \includegraphics[width=\textwidth]{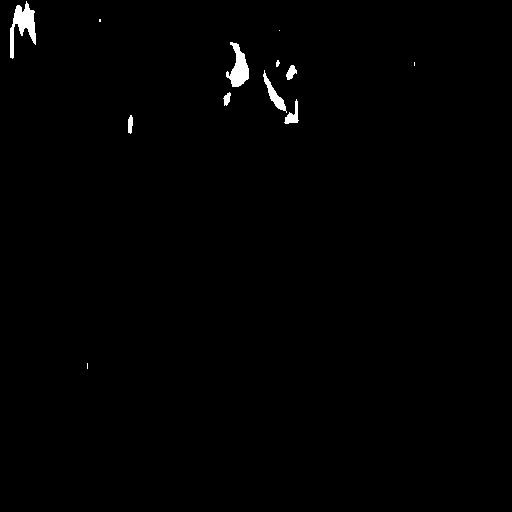}
     \end{subfigure}
     \hfill
     \begin{subfigure}[b]{0.16\textwidth}
         \centering
            \includegraphics[width=\textwidth]{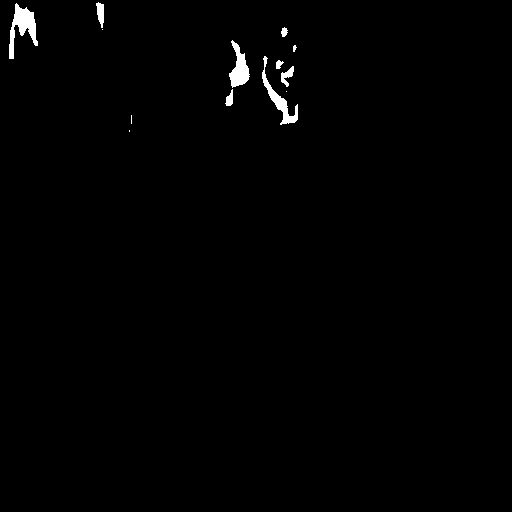}
     \end{subfigure}
     \hfill
     \begin{subfigure}[b]{0.16\textwidth}
         \centering
            \includegraphics[width=\textwidth]{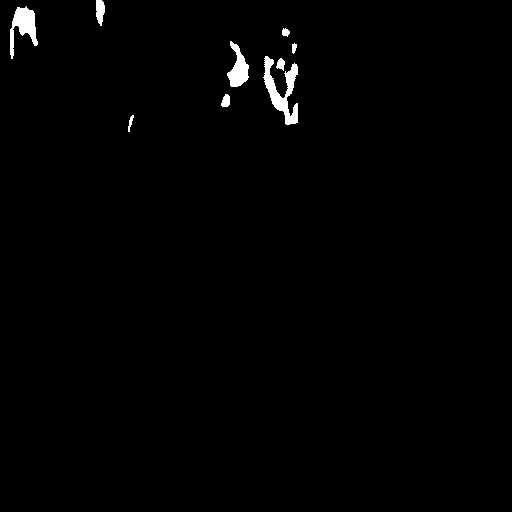}
     \end{subfigure}
     \\
     \begin{subfigure}[b]{\textwidth}
     \end{subfigure}
     \\
     \begin{subfigure}[b]{0.16\textwidth}
         \centering
            \includegraphics[width=\textwidth]{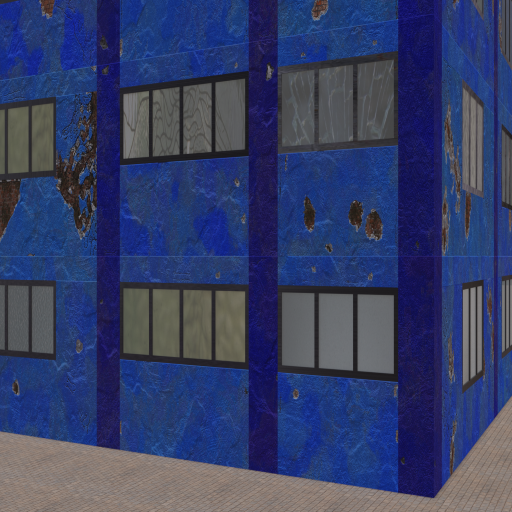}
         \caption{Input image}
     \end{subfigure}
     \hfill
     \begin{subfigure}[b]{0.16\textwidth}
         \centering
            \includegraphics[width=\textwidth]{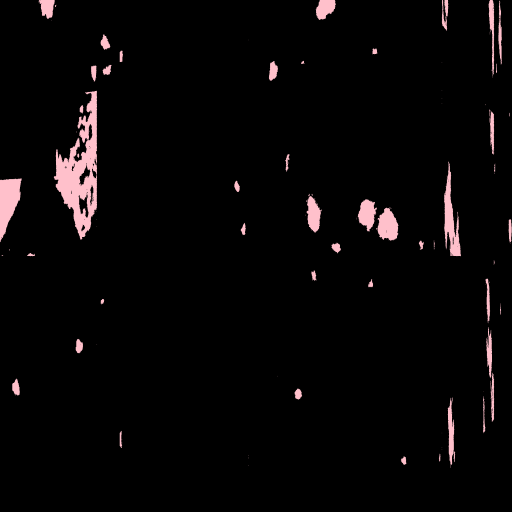}
         \caption{Ground truth}
     \end{subfigure}
     \hfill
     \begin{subfigure}[b]{0.16\textwidth}
         \centering
            \includegraphics[width=\textwidth]{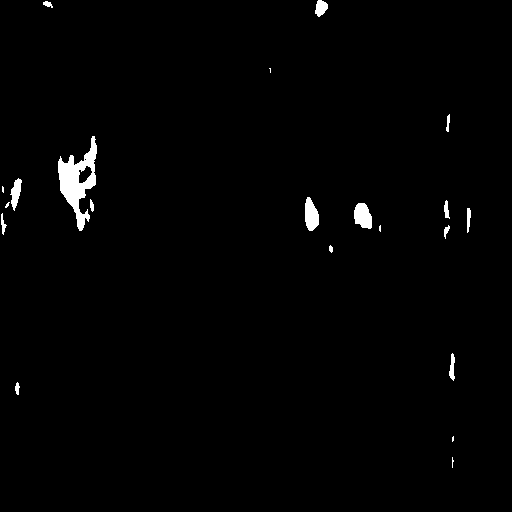}
         \caption{U-net}
     \end{subfigure}
     \hfill
     \begin{subfigure}[b]{0.16\textwidth}
         \centering
            \includegraphics[width=\textwidth]{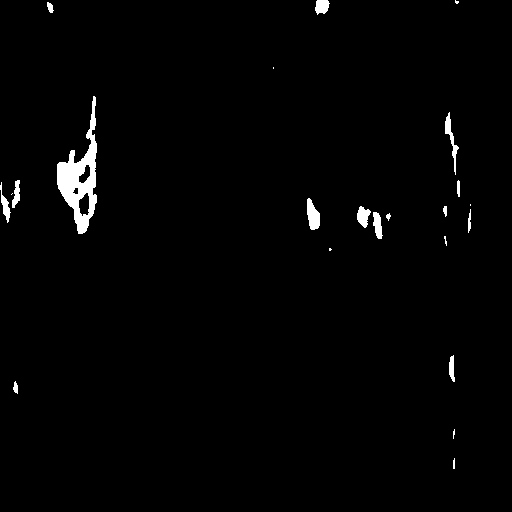}
         \caption{Adv. attn. U-net}
     \end{subfigure}
     \hfill
     \begin{subfigure}[b]{0.16\textwidth}
         \centering
            \includegraphics[width=\textwidth]{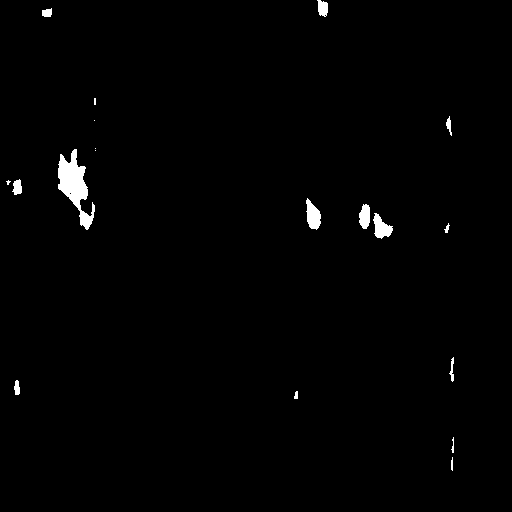}
         \caption{U-net + D-6}
     \end{subfigure}
     \hfill
     \begin{subfigure}[b]{0.16\textwidth}
         \centering
            \includegraphics[width=\textwidth]{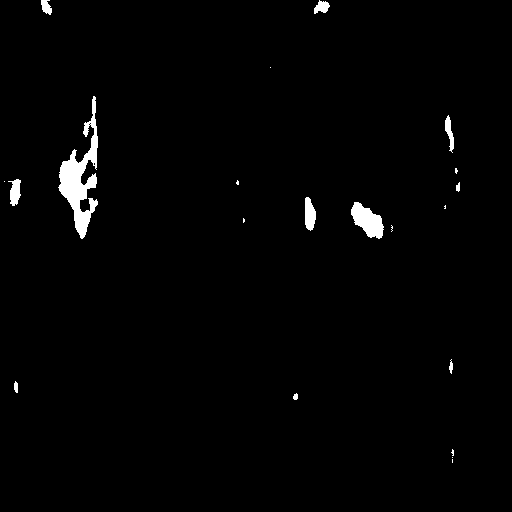}
            \caption{Com. network}
     \end{subfigure}
    \caption{Data and semantic segmentation results for the on-site inspection}
    \label{fig:imp2}
\end{figure*}

The evaluation of the segmentation quality is executed through two steps. The PA and mIoU of the best ten results are firstly calculated (see Tab. \ref{tab:cvvald1}). After that, the entire 50 segmentation images are evaluated by the two same metrics (see Tab. \ref{tab:cvvald1}). The original images corresponding to the selected results have several common features, such as larger areas, distinct edges, and higher color contrast between the spall and other objects (see Fig. \ref{fig:imp2}).

As the values in Tab. \ref{tab:cvvald1} indicate, three enhanced architectures materialise the prominent improvement in semantic segmentation with different significance. The advanced attention U-net results in a better evaluation in 50 images than the selected ten. The U-net attached by the D-6 presents a robust capability of semantic segmentation, the quality of input images on that is much less than the advanced attention U-net. Though the D-6 U-net provides a higher PA and mIoU than the advanced attention U-net in segmenting the ten selected images, both architectures show comparable performance in facing a more generic dataset. The combined network shows stable results, which are also better than the U-net. Nevertheless, the results appear to have adverse effects from combining the attention mechanism and a discriminator in a U-net because the evaluations of the combined network are apparently lower than the other two enhanced U-nets. 

\begin{table}[htbp]
\caption{Result evaluation of the 10 and 50 selected images in validation}
\label{tab:cvvald1}
\centering
\begin{threeparttable}
\begin{tabular}{lcccccc}
\toprule
 \textbf{10 images} & \textbf{PA\tnote{$^{\rm a}$}} (\%) & \textbf{$\Delta$}\tnote{$^{\rm b}$} (\%) & \textbf{mIoU}\tnote{$^{\rm c}$} (\%) & \textbf{$\Delta$} (\%)  \\
 \midrule
U-net & 96.06 & - & 75.29 & -  \\
Adv. attn. U-net \tnote{$^{\rm d}$} & 97.19 & +1.16 & 81.15 & +7.78 \\
U-net + D-6 & \textbf{97.38} & \textbf{+1.37} & \textbf{82.50} & \textbf{+9.58} \\
Com. network \tnote{$^{\rm e}$} & 96.98 & +0.96 & 80.17 & +6.48 \\
\bottomrule
\end{tabular}
\quad
\begin{tabular}{lcccccc}
\toprule
 \textbf{50 images} & \textbf{PA\tnote{$^{\rm a}$}} (\%) & \textbf{$\Delta$}\tnote{$^{\rm b}$} (\%) & \textbf{mIoU}\tnote{$^{\rm c}$} (\%) & \textbf{$\Delta$} (\%)  \\
 \midrule
U-net & 96.02 & - & 68.64 & -  \\
Adv. attn. U-net \tnote{$^{\rm d}$}  & \textbf{97.98} & \textbf{+2.04} & \textbf{74.97} & \textbf{+9.22} \\
U-net + D-6 & \textbf{98.02} & \textbf{+2.06} & \textbf{74.80} & \textbf{+8.97} \\
Com. network \tnote{$^{\rm e}$} & 97.80 & +1.85 & 73.04 & +6.41 \\
\bottomrule
\end{tabular}

\begin{tablenotes}
\item[$^{\rm a}$] PA: Pixel Accuracy
\item[$^{\rm b}$] $\Delta$: the increment compared with the results from the U-net
\item[$^{\rm c}$] mIoU: the mean Intersection over Union
\item[$^{\rm d}$] Adv. attn. U-net: the advanced attention U-net
\item[$^{\rm e}$] Com. network: the network combined with the advanced attention U-net and the discriminator D-6
\end{tablenotes}
\end{threeparttable}
\end{table}

The semantic segmentation results reveal another angle of observation. Fig. \ref{fig:imp2} depicts that the strength of the attention mechanism is prone to detect the edges of an object. The combined network performs with a weaker edge identification, and the U-net + D-6 is the worst one. This observation seems does not endorse the calculated metrics. However, as a matter of fact, the metrics, PA and mIoU, are widely applied for a more comprehensive evaluation to examine the exact locations of each pixel area. Visual observation is exclusively able to discover few aspects in result assessment.

\section{Discussion}\label{sec6}

\begin{figure*}[ht]
     \centering
        \includegraphics[width=0.7\textwidth]{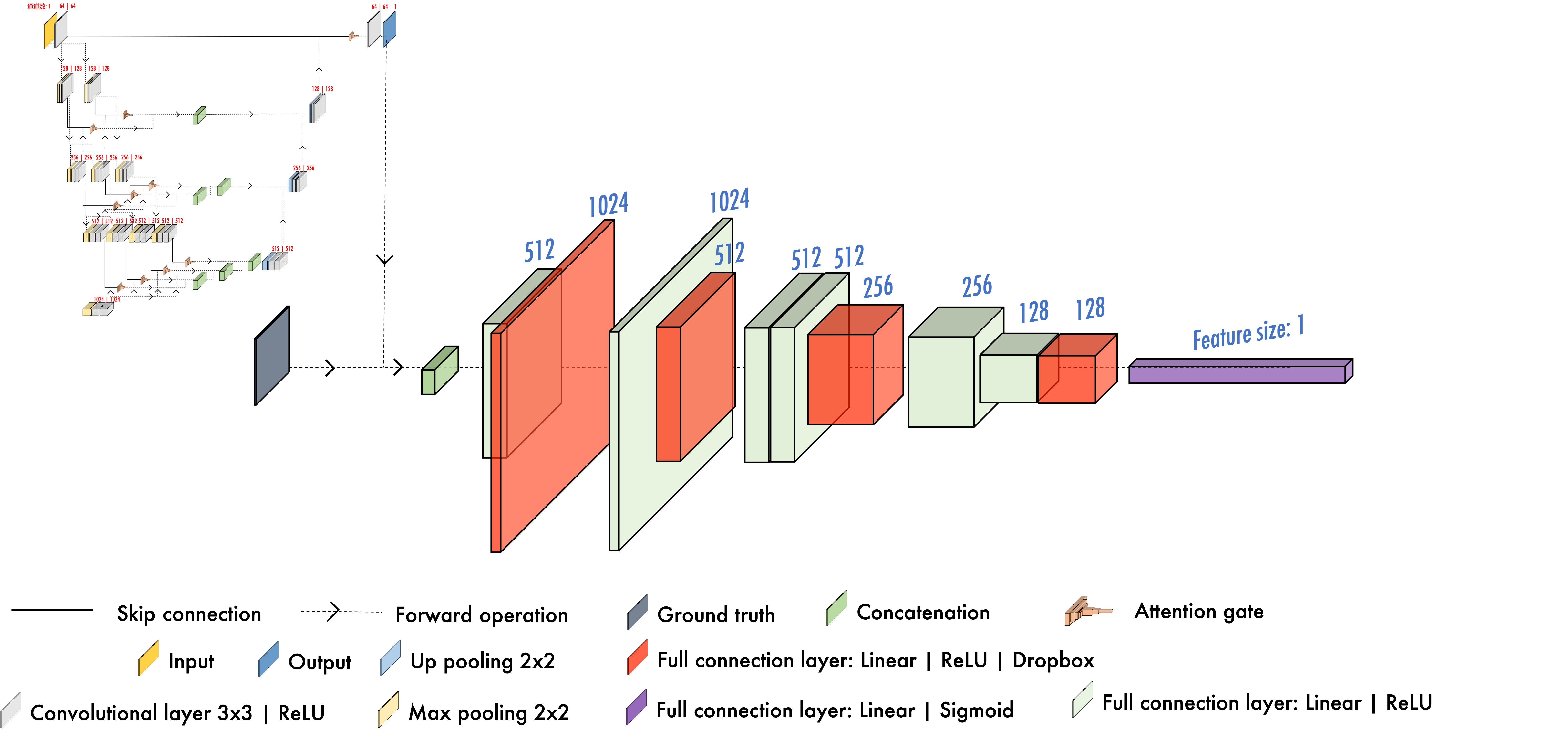}
     \caption{Com. network 2}
     \label{fig:ComNet2}
\end{figure*}

The attention mechanism and the discriminator are supposed to be two individual components in the frame of a GAN. The former extracts the features from the inputs in a generator's different stages; the latter evaluates the generator's outputs and feeds back for parameter adjustment during training. Incorporating the attention mechanism into a GAN is primitively expected to combine these two components' strengths. On the contrary, Tab. \ref{tab:cvvald1} proves this kind of combination does not function as the expectation. The segmentation quality of the combined network is worse than the other two enhanced U-nets rather than reaches the best values. To the author's best knowledge, this inharmonious orchestration between the attention mechanism and a discriminator was the very first to be discovered and discussed.

In order to investigate whether the inharmonious orchestration is a contingency or not, another group of architectures is chosen from a comparative study, i.e., the full attention U-net, the D-4 U-net, and their combination (see Fig \ref{fig:ComNet2}). These three models are trained with the identical training configuration to the previous group in the lighet weight test, and tested by the same dataset with 50 images. According to the assessment in Tab. \ref{tab:cvvald3}, this combination does work better than the other two U-nets. However, none of the tested architecture delivers more optimal values of PA and mIoU.

\begin{table}[htbp]
\caption{Result evaluation of the 50 images for the comparative group}
\label{tab:cvvald3}
\centering
\begin{threeparttable}
\begin{tabular}{lcccccc}
\toprule
 & \textbf{PA\tnote{$^{\rm a}$}} (\%) & \textbf{$\Delta$}\tnote{$^{\rm b}$} (\%) & \textbf{mIoU}\tnote{$^{\rm c}$} (\%) & \textbf{$\Delta$} (\%)  \\
 \midrule
U-net & 96.02 & - & 68.64 & -  \\
Full attn. U-net  \tnote{$^{\rm d}$} & 97.33 & +1.36 & 67.87 & +1.12 \\
U-net + D-4 & 97.45 & +1.49 & 68.98 & +0.50 \\
Com. network 2 \tnote{$^{\rm e}$} & \textbf{97.55} & \textbf{+1.59} & \textbf{70.14} & \textbf{+2.19} \\
\bottomrule
\end{tabular}

\begin{tablenotes}
\item[$^{\rm a}$] PA: Pixel Accuracy
\item[$^{\rm b}$] $\Delta$: the increment compared with the results from the U-net
\item[$^{\rm c}$] mIoU: the mean Intersection over Union
\item[$^{\rm d}$] Full attn. U-net: the full attention U-net
\item[$^{\rm e}$] Com. network 2: the network combined with the full attention U-net and the discriminator D-4V
\end{tablenotes}
\end{threeparttable}
\end{table}

After all, this test manifests that the collaboration of the attention mechanism and a discriminator could negatively impact semantic segmentation, but exclusively in certain circumstances. One possible explanation could be that the feedback information transmitted from the discriminator to the generator interferes with the attention operation on the decoder's side. The full attention mechanism is not or barely influenced because most of the attention gates are distributed on the decoder's side. This analysis is partially endorsed by the GAN model developed by \cite{zhang_semi-supervised_2021}. That proposed GAN is integrated with the attention mechanism as well. However, the attention block is placed in front of the U-net. In other words, the data fed to the inputs have been previously attended outside of the U-net. The mathematical analysis could be agnostic and heterogeneous so that further study should focus on this phenomenon.

\section{Conclusion}\label{sec7}

This paper presents a comparative study for semantic segmentation focusing on on-site inspection. Inspired by the attention mechanism and the design of the U-net, two novel architectures integrated with attention gates are proposed, i.e., the advanced attention U-net and the full attention U-net. Furthermore, four novel architectures of discriminators in U-net based GANs are proposed as well, namely the D-4 U-net, the D-6 U-net, the D-4V U-net, and the D-5V U-net. A lightweight experiment is executed to compare these enhanced U-net architectures in two groups. The U-net, the attention U-net and the two brand-newly proposed U-nets with attention mechanisms are marked as the first group, in which the advanced attention U-net achieves the best results of semantic segmentation. Four newly proposed GANs are compared with the U-net in the second group. Consequently, the D-6 U-net shows the highest segmentation accuracy. Hence, the advanced U-net, the D-6 U-net, and the U-net combined of both are selected for the further implementation, where the graphical data acquired from a pseudo on-site inspection. The results prove the improved capability of the tested architectures in semantic segmentation. PA and mIoU are calculated to designate the strengths of the advanced U-net and the D-6 U-net in segmenting ten selected images and the whole dataset. 

Meanwhile, the adverse effects occasioned by the combination of a discriminator and attention mechanism are discovered.  For further investigation, another group is trained and tested in the experiment, including the full attention U-net, the D-4 U-net, and the U-net combined with both. These test results demonstrate positive combination effects and thus stand on the opposite of the previous test. The comparison with all the groups ascribes this phenomenon to the different distribution of attention gates in the proposed enhanced U-nets.

It is to be noticed that the inherent shortcoming of the U-net lies in segmenting slim objects. Though the spall is well recognized in this research work, semantic segmentation of slim damage such as cracks can not be easily segmented. Therefore, other strategies, e.g., featured pyramid network and residual operation, could be integrated in the future to promote the semantic segmentation quality of the proposed CNNs.

\section*{Acknowledgments}
The authors would like to acknowledge the organization committee of The 2nd International Competition for Structural Health Monitoring (IC-SHM, 2021) that shared the graphical data for training and test, and the hardware support from Microintelligenc Inc. 

\section*{Conflict of interest}
The authors declare no potential conflict of interests.

\balance
\bibliography{wileyNJD-APA}%

\begin{thebibliography}{}

\bibitem [\protect \citeauthoryear {%
Abadi%
\ \protect \BOthers {.}}{%
Abadi%
\ \protect \BOthers {.}}{%
{\protect \APACyear {2015}}%
}]{%
tensorflow2015-whitepaper}
\APACinsertmetastar {%
tensorflow2015-whitepaper}%
\begin{APACrefauthors}%
Abadi, M.%
, Agarwal, A.%
, Barham, P.%
, Brevdo, E.%
, Chen, Z.%
, Citro, C.%
\BDBL {}Zheng, X.%
\end{APACrefauthors}%
\unskip\
\newblock
\APACrefYearMonthDay{2015}{}{}.
\newblock
\APACrefbtitle {{TensorFlow}: Large-Scale Machine Learning on Heterogeneous
  Systems.} {{TensorFlow}: Large-scale machine learning on heterogeneous
  systems.}
\newblock
\begin{APACrefURL} \url{http://tensorflow.org/} \end{APACrefURL}
\newblock
\APACrefnote{Software available from tensorflow.org}
\PrintBackRefs{\CurrentBib}

\bibitem [\protect \citeauthoryear {%
Bahdanau%
, Cho%
\BCBL {}\ \BBA {} Bengio%
}{%
Bahdanau%
\ \protect \BOthers {.}}{%
{\protect \APACyear {2015}}%
}]{%
bahdanau_neural_2015}
\APACinsertmetastar {%
bahdanau_neural_2015}%
\begin{APACrefauthors}%
Bahdanau, D.%
, Cho, K\BPBI H.%
\BCBL {}\ \BBA {} Bengio, Y.%
\end{APACrefauthors}%
\unskip\
\newblock
\APACrefYearMonthDay{2015}{}{}.
\newblock
{\BBOQ}\APACrefatitle {Neural machine translation by jointly learning to align
  and translate} {Neural machine translation by jointly learning to align and
  translate}.{\BBCQ}
\newblock
\BIn{} \APACrefbtitle {The proceedings of 3rd {International} {Conference} on
  {Learning} {Representations}.} {The proceedings of 3rd {International}
  {Conference} on {Learning} {Representations}.}
\newblock
\APACaddressPublisher{San Diego}{}.
\newblock
\begin{APACrefURL}
  [{2021-09-17}]\url{http://www.scopus.com/inward/record.url?scp=85062889504&partnerID=8YFLogxK}
  \end{APACrefURL}
\PrintBackRefs{\CurrentBib}

\bibitem [\protect \citeauthoryear {%
Beek%
, Reinders%
, Sankur%
\BCBL {}\ \BBA {} Lubbe%
}{%
Beek%
\ \protect \BOthers {.}}{%
{\protect \APACyear {1992}}%
}]{%
beek_semantic_1992}
\APACinsertmetastar {%
beek_semantic_1992}%
\begin{APACrefauthors}%
Beek, P\BPBI J\BPBI L\BPBI v.%
, Reinders, M\BPBI J\BPBI T.%
, Sankur, B.%
\BCBL {}\ \BBA {} Lubbe, J\BPBI C\BPBI A\BPBI v\BPBI d.%
\end{APACrefauthors}%
\unskip\
\newblock
\APACrefYearMonthDay{1992}{}{}.
\newblock
{\BBOQ}\APACrefatitle {Semantic segmentation of videophone image sequences}
  {Semantic segmentation of videophone image sequences}.{\BBCQ}
\newblock
\BIn{} \APACrefbtitle {Visual {Communications} and {Image} {Processing} '92}
  {Visual {Communications} and {Image} {Processing} '92}\ (\BVOL\ 1818, \BPGS\
  1182--1193).
\newblock
\APACaddressPublisher{}{International Society for Optics and Photonics}.
\newblock
\begin{APACrefDOI} \doi{10.1117/12.131389} \end{APACrefDOI}
\PrintBackRefs{\CurrentBib}

\bibitem [\protect \citeauthoryear {%
Chen%
\ \protect \BOthers {.}}{%
Chen%
\ \protect \BOthers {.}}{%
{\protect \APACyear {2020}}%
}]{%
chen_image-based_2020}
\APACinsertmetastar {%
chen_image-based_2020}%
\begin{APACrefauthors}%
Chen, J.%
, Zhang, D.%
, Huang, H.%
, Shadabfar, M.%
, Zhou, M.%
\BCBL {}\ \BBA {} Yang, T.%
\end{APACrefauthors}%
\unskip\
\newblock
\APACrefYearMonthDay{2020}{}{}.
\newblock
{\BBOQ}\APACrefatitle {Image-based segmentation and quantification of weak
  interlayers in rock tunnel face via deep learning} {Image-based segmentation
  and quantification of weak interlayers in rock tunnel face via deep
  learning}.{\BBCQ}
\newblock
\APACjournalVolNumPages{Automation in Construction}{120}{}{103371}.
\newblock
\begin{APACrefDOI} \doi{10.1016/j.autcon.2020.103371} \end{APACrefDOI}
\PrintBackRefs{\CurrentBib}

\bibitem [\protect \citeauthoryear {%
Chollet%
\ \protect \BOthers {.}}{%
Chollet%
\ \protect \BOthers {.}}{%
{\protect \APACyear {2015}}%
}]{%
chollet2015keras}
\APACinsertmetastar {%
chollet2015keras}%
\begin{APACrefauthors}%
Chollet, F.%
\BCBT {}\ \BOthersPeriod {.}
\end{APACrefauthors}%
\unskip\
\newblock
\APACrefYearMonthDay{2015}{}{}.
\newblock
\APACrefbtitle {Keras.} {Keras.}
\newblock
\APAChowpublished {\url{https://keras.io}}.
\PrintBackRefs{\CurrentBib}

\bibitem [\protect \citeauthoryear {%
Csurka%
\ \BBA {} Perronnin%
}{%
Csurka%
\ \BBA {} Perronnin%
}{%
{\protect \APACyear {2011}}%
}]{%
csurka_efficient_2011}
\APACinsertmetastar {%
csurka_efficient_2011}%
\begin{APACrefauthors}%
Csurka, G.%
\BCBT {}\ \BBA {} Perronnin, F.%
\end{APACrefauthors}%
\unskip\
\newblock
\APACrefYearMonthDay{2011}{}{}.
\newblock
{\BBOQ}\APACrefatitle {An efficient approach to semantic segmentation} {An
  efficient approach to semantic segmentation}.{\BBCQ}
\newblock
\APACjournalVolNumPages{International Journal of Computer
  Vision}{95}{2}{198--212}.
\newblock
\begin{APACrefDOI} \doi{10.1007/s11263-010-0344-8} \end{APACrefDOI}
\PrintBackRefs{\CurrentBib}

\bibitem [\protect \citeauthoryear {%
Czerniawski%
\ \BBA {} Leite%
}{%
Czerniawski%
\ \BBA {} Leite%
}{%
{\protect \APACyear {2020}}%
}]{%
czerniawski_automated_2020}
\APACinsertmetastar {%
czerniawski_automated_2020}%
\begin{APACrefauthors}%
Czerniawski, T.%
\BCBT {}\ \BBA {} Leite, F.%
\end{APACrefauthors}%
\unskip\
\newblock
\APACrefYearMonthDay{2020}{}{}.
\newblock
{\BBOQ}\APACrefatitle {Automated segmentation of {RGB}-{D} images into a
  comprehensive set of building components using deep learning} {Automated
  segmentation of {RGB}-{D} images into a comprehensive set of building
  components using deep learning}.{\BBCQ}
\newblock
\APACjournalVolNumPages{Advanced Engineering Informatics}{45}{}{101131}.
\newblock
\begin{APACrefDOI} \doi{10.1016/j.aei.2020.101131} \end{APACrefDOI}
\PrintBackRefs{\CurrentBib}

\bibitem [\protect \citeauthoryear {%
Dais%
, Bal%
, Smyrou%
\BCBL {}\ \BBA {} Sarhosis%
}{%
Dais%
\ \protect \BOthers {.}}{%
{\protect \APACyear {2021}}%
}]{%
dais_automatic_2021}
\APACinsertmetastar {%
dais_automatic_2021}%
\begin{APACrefauthors}%
Dais, D.%
, Bal, Ä\BPBI E.%
, Smyrou, E.%
\BCBL {}\ \BBA {} Sarhosis, V.%
\end{APACrefauthors}%
\unskip\
\newblock
\APACrefYearMonthDay{2021}{}{}.
\newblock
{\BBOQ}\APACrefatitle {Automatic crack classification and segmentation on
  masonry surfaces using convolutional neural networks and transfer learning}
  {Automatic crack classification and segmentation on masonry surfaces using
  convolutional neural networks and transfer learning}.{\BBCQ}
\newblock
\APACjournalVolNumPages{Automation in Construction}{125}{}{103606}.
\newblock
\begin{APACrefDOI} \doi{10.1016/j.autcon.2021.103606} \end{APACrefDOI}
\PrintBackRefs{\CurrentBib}

\bibitem [\protect \citeauthoryear {%
Decourt%
\ \BBA {} Duong%
}{%
Decourt%
\ \BBA {} Duong%
}{%
{\protect \APACyear {2020}}%
}]{%
decourt_semi-supervised_2020}
\APACinsertmetastar {%
decourt_semi-supervised_2020}%
\begin{APACrefauthors}%
Decourt, C.%
\BCBT {}\ \BBA {} Duong, L.%
\end{APACrefauthors}%
\unskip\
\newblock
\APACrefYearMonthDay{2020}{}{}.
\newblock
{\BBOQ}\APACrefatitle {Semi-supervised generative adversarial networks for the
  segmentation of the left ventricle in pediatric {MRI}} {Semi-supervised
  generative adversarial networks for the segmentation of the left ventricle in
  pediatric {MRI}}.{\BBCQ}
\newblock
\APACjournalVolNumPages{Computers in Biology and Medicine}{123}{}{103884}.
\newblock
\begin{APACrefDOI} \doi{10.1016/j.compbiomed.2020.103884} \end{APACrefDOI}
\PrintBackRefs{\CurrentBib}

\bibitem [\protect \citeauthoryear {%
Ghassemi%
, Shoeibi%
\BCBL {}\ \BBA {} Rouhani%
}{%
Ghassemi%
\ \protect \BOthers {.}}{%
{\protect \APACyear {2020}}%
}]{%
ghassemi_deep_2020}
\APACinsertmetastar {%
ghassemi_deep_2020}%
\begin{APACrefauthors}%
Ghassemi, N.%
, Shoeibi, A.%
\BCBL {}\ \BBA {} Rouhani, M.%
\end{APACrefauthors}%
\unskip\
\newblock
\APACrefYearMonthDay{2020}{}{}.
\newblock
{\BBOQ}\APACrefatitle {Deep neural network with generative adversarial networks
  pre-training for brain tumor classification based on {MR} images} {Deep
  neural network with generative adversarial networks pre-training for brain
  tumor classification based on {MR} images}.{\BBCQ}
\newblock
\APACjournalVolNumPages{Biomedical Signal Processing and
  Control}{57}{}{101678}.
\newblock
\begin{APACrefDOI} \doi{10.1016/j.bspc.2019.101678} \end{APACrefDOI}
\PrintBackRefs{\CurrentBib}

\bibitem [\protect \citeauthoryear {%
Girshick%
}{%
Girshick%
}{%
{\protect \APACyear {2015}}%
}]{%
girshick_fast_2015}
\APACinsertmetastar {%
girshick_fast_2015}%
\begin{APACrefauthors}%
Girshick, R.%
\end{APACrefauthors}%
\unskip\
\newblock
\APACrefYearMonthDay{2015}{}{}.
\newblock
{\BBOQ}\APACrefatitle {Fast {R}-{CNN}} {Fast {R}-{CNN}}.{\BBCQ}
\newblock
\APACjournalVolNumPages{arXiv:1504.08083 [cs]}{}{}{}.
\PrintBackRefs{\CurrentBib}

\bibitem [\protect \citeauthoryear {%
Goodfellow%
\ \protect \BOthers {.}}{%
Goodfellow%
\ \protect \BOthers {.}}{%
{\protect \APACyear {2014}}%
}]{%
Goodfellow2014GenerativeAN}
\APACinsertmetastar {%
Goodfellow2014GenerativeAN}%
\begin{APACrefauthors}%
Goodfellow, I\BPBI J.%
, Pouget-Abadie, J.%
, Mirza, M.%
, Xu, B.%
, Warde-Farley, D.%
, Ozair, S.%
\BDBL {}Bengio, Y.%
\end{APACrefauthors}%
\unskip\
\newblock
\APACrefYearMonthDay{2014}{}{}.
\newblock
{\BBOQ}\APACrefatitle {Generative Adversarial Networks} {Generative adversarial
  networks}.{\BBCQ}
\newblock
\APACjournalVolNumPages{ArXiv}{abs/1406.2661}{}{}.
\PrintBackRefs{\CurrentBib}

\bibitem [\protect \citeauthoryear {%
Hang%
, Li%
, Liu%
, Ghamisi%
\BCBL {}\ \BBA {} Bhattacharyya%
}{%
Hang%
\ \protect \BOthers {.}}{%
{\protect \APACyear {2021}}%
}]{%
hang_hyperspectral_2021}
\APACinsertmetastar {%
hang_hyperspectral_2021}%
\begin{APACrefauthors}%
Hang, R.%
, Li, Z.%
, Liu, Q.%
, Ghamisi, P.%
\BCBL {}\ \BBA {} Bhattacharyya, S\BPBI S.%
\end{APACrefauthors}%
\unskip\
\newblock
\APACrefYearMonthDay{2021}{}{}.
\newblock
{\BBOQ}\APACrefatitle {Hyperspectral {image} {classification} {with}
  {attention}-{aided} {CNNs}} {Hyperspectral {image} {classification} {with}
  {attention}-{aided} {CNNs}}.{\BBCQ}
\newblock
\APACjournalVolNumPages{IEEE Transactions on Geoscience and Remote
  Sensing}{59}{3}{2281--2293}.
\newblock
\APACrefnote{Conference Name: IEEE Transactions on Geoscience and Remote
  Sensing}
\newblock
\begin{APACrefDOI} \doi{10.1109/TGRS.2020.3007921} \end{APACrefDOI}
\PrintBackRefs{\CurrentBib}

\bibitem [\protect \citeauthoryear {%
He%
, Gkioxari%
, Dollár%
\BCBL {}\ \BBA {} Girshick%
}{%
He%
\ \protect \BOthers {.}}{%
{\protect \APACyear {2018}}%
}]{%
he_mask_2018}
\APACinsertmetastar {%
he_mask_2018}%
\begin{APACrefauthors}%
He, K.%
, Gkioxari, G.%
, Dollár, P.%
\BCBL {}\ \BBA {} Girshick, R.%
\end{APACrefauthors}%
\unskip\
\newblock
\APACrefYearMonthDay{2018}{}{}.
\newblock
{\BBOQ}\APACrefatitle {Mask {R}-{CNN}} {Mask {R}-{CNN}}.{\BBCQ}
\newblock
\APACjournalVolNumPages{arXiv:1703.06870 [cs]}{}{}{}.
\PrintBackRefs{\CurrentBib}

\bibitem [\protect \citeauthoryear {%
He%
, Zhang%
, Ren%
\BCBL {}\ \BBA {} Sun%
}{%
He%
\ \protect \BOthers {.}}{%
{\protect \APACyear {2015}}%
}]{%
he_deep_2015}
\APACinsertmetastar {%
he_deep_2015}%
\begin{APACrefauthors}%
He, K.%
, Zhang, X.%
, Ren, S.%
\BCBL {}\ \BBA {} Sun, J.%
\end{APACrefauthors}%
\unskip\
\newblock
\APACrefYearMonthDay{2015}{}{}.
\newblock
{\BBOQ}\APACrefatitle {Deep residual learning for image recognition} {Deep
  residual learning for image recognition}.{\BBCQ}
\newblock
\APACjournalVolNumPages{arXiv:1512.03385 [cs]}{}{}{}.
\PrintBackRefs{\CurrentBib}

\bibitem [\protect \citeauthoryear {%
He%
, Zhang%
, Ren%
\BCBL {}\ \BBA {} Sun%
}{%
He%
\ \protect \BOthers {.}}{%
{\protect \APACyear {2016}}%
}]{%
He2016DeepRL}
\APACinsertmetastar {%
He2016DeepRL}%
\begin{APACrefauthors}%
He, K.%
, Zhang, X.%
, Ren, S.%
\BCBL {}\ \BBA {} Sun, J.%
\end{APACrefauthors}%
\unskip\
\newblock
\APACrefYearMonthDay{2016}{}{}.
\newblock
{\BBOQ}\APACrefatitle {Deep Residual Learning for Image Recognition} {Deep
  residual learning for image recognition}.{\BBCQ}
\newblock
\APACjournalVolNumPages{2016 IEEE Conference on Computer Vision and Pattern
  Recognition (CVPR)}{}{}{770-778}.
\newblock
\begin{APACrefDOI} \doi{10.1109/cvpr.2016.90} \end{APACrefDOI}
\PrintBackRefs{\CurrentBib}

\bibitem [\protect \citeauthoryear {%
H.~Huang%
\ \protect \BOthers {.}}{%
H.~Huang%
\ \protect \BOthers {.}}{%
{\protect \APACyear {2020}}%
}]{%
huang_unet_2020}
\APACinsertmetastar {%
huang_unet_2020}%
\begin{APACrefauthors}%
Huang, H.%
, Lin, L.%
, Tong, R.%
, Hu, H.%
, Zhang, Q.%
, Iwamoto, Y.%
\BDBL {}Wu, J.%
\end{APACrefauthors}%
\unskip\
\newblock
\APACrefYearMonthDay{2020}{}{}.
\newblock
{\BBOQ}\APACrefatitle {{UNet} 3+: {A} full-scale connected {UNet} for medical
  image segmentation} {{UNet} 3+: {A} full-scale connected {UNet} for medical
  image segmentation}.{\BBCQ}
\newblock
\APACjournalVolNumPages{arXiv:2004.08790 [cs, eess]}{}{}{}.
\PrintBackRefs{\CurrentBib}

\bibitem [\protect \citeauthoryear {%
J.~Huang%
, Liu%
, Tang%
\BCBL {}\ \BBA {} Zhang%
}{%
J.~Huang%
\ \protect \BOthers {.}}{%
{\protect \APACyear {2021}}%
}]{%
huang_object-level_2021}
\APACinsertmetastar {%
huang_object-level_2021}%
\begin{APACrefauthors}%
Huang, J.%
, Liu, S.%
, Tang, Y.%
\BCBL {}\ \BBA {} Zhang, X.%
\end{APACrefauthors}%
\unskip\
\newblock
\APACrefYearMonthDay{2021}{}{}.
\newblock
{\BBOQ}\APACrefatitle {Object-{Level} {Remote} {Sensing} {Image} {Augmentation}
  {Using} {U}-{Net}-{Based} {Generative} {Adversarial} {Networks}}
  {Object-{Level} {Remote} {Sensing} {Image} {Augmentation} {Using}
  {U}-{Net}-{Based} {Generative} {Adversarial} {Networks}}.{\BBCQ}
\newblock
\APACjournalVolNumPages{Wireless Communications and Mobile
  Computing}{2021}{}{e1230279}.
\newblock
\begin{APACrefDOI} \doi{10.1155/2021/1230279} \end{APACrefDOI}
\PrintBackRefs{\CurrentBib}

\bibitem [\protect \citeauthoryear {%
Inoguchi%
, Tamura%
\BCBL {}\ \BBA {} Hamamoto%
}{%
Inoguchi%
\ \protect \BOthers {.}}{%
{\protect \APACyear {2019}}%
}]{%
inoguchi_establishment_2019}
\APACinsertmetastar {%
inoguchi_establishment_2019}%
\begin{APACrefauthors}%
Inoguchi, M.%
, Tamura, K.%
\BCBL {}\ \BBA {} Hamamoto, R.%
\end{APACrefauthors}%
\unskip\
\newblock
\APACrefYearMonthDay{2019}{}{}.
\newblock
{\BBOQ}\APACrefatitle {Establishment of work-flow for roof damage detection
  utilizing drones, human and {AI} based on human-in-the-loop framework}
  {Establishment of work-flow for roof damage detection utilizing drones, human
  and {AI} based on human-in-the-loop framework}.{\BBCQ}
\newblock
\BIn{} \APACrefbtitle {2019 {IEEE} {International} {Conference} on {Big} {Data}
  ({Big} {Data})} {2019 {IEEE} {International} {Conference} on {Big} {Data}
  ({Big} {Data})}\ (\BPGS\ 4618--4623).
\newblock
\begin{APACrefDOI} \doi{10.1109/BigData47090.2019.9006211} \end{APACrefDOI}
\PrintBackRefs{\CurrentBib}

\bibitem [\protect \citeauthoryear {%
Krizhevsky%
, Sutskever%
\BCBL {}\ \BBA {} Hinton%
}{%
Krizhevsky%
\ \protect \BOthers {.}}{%
{\protect \APACyear {2017}}%
}]{%
krizhevsky_imagenet_2017}
\APACinsertmetastar {%
krizhevsky_imagenet_2017}%
\begin{APACrefauthors}%
Krizhevsky, A.%
, Sutskever, I.%
\BCBL {}\ \BBA {} Hinton, G\BPBI E.%
\end{APACrefauthors}%
\unskip\
\newblock
\APACrefYearMonthDay{2017}{}{}.
\newblock
{\BBOQ}\APACrefatitle {{ImageNet} classification with deep convolutional neural
  networks} {{ImageNet} classification with deep convolutional neural
  networks}.{\BBCQ}
\newblock
\APACjournalVolNumPages{Communications of the ACM}{60}{6}{84--90}.
\newblock
\begin{APACrefDOI} \doi{10.1145/3065386} \end{APACrefDOI}
\PrintBackRefs{\CurrentBib}

\bibitem [\protect \citeauthoryear {%
Lecun%
, Bottou%
, Bengio%
\BCBL {}\ \BBA {} Haffner%
}{%
Lecun%
\ \protect \BOthers {.}}{%
{\protect \APACyear {1998}}%
}]{%
lecun_gradient-based_1998}
\APACinsertmetastar {%
lecun_gradient-based_1998}%
\begin{APACrefauthors}%
Lecun, Y.%
, Bottou, L.%
, Bengio, Y.%
\BCBL {}\ \BBA {} Haffner, P.%
\end{APACrefauthors}%
\unskip\
\newblock
\APACrefYearMonthDay{1998}{}{}.
\newblock
{\BBOQ}\APACrefatitle {Gradient-based learning applied to document recognition}
  {Gradient-based learning applied to document recognition}.{\BBCQ}
\newblock
\APACjournalVolNumPages{Proceedings of the IEEE}{86}{11}{2278--2324}.
\newblock
\begin{APACrefDOI} \doi{10.1109/5.726791} \end{APACrefDOI}
\PrintBackRefs{\CurrentBib}

\bibitem [\protect \citeauthoryear {%
LeCun%
, Cortes%
\BCBL {}\ \BBA {} Burges%
}{%
LeCun%
\ \protect \BOthers {.}}{%
{\protect \APACyear {1998}}%
}]{%
lecun_mnist_1998}
\APACinsertmetastar {%
lecun_mnist_1998}%
\begin{APACrefauthors}%
LeCun, Y.%
, Cortes, C.%
\BCBL {}\ \BBA {} Burges, C\BPBI J.%
\end{APACrefauthors}%
\unskip\
\newblock
\APACrefYearMonthDay{1998}{}{}.
\newblock
\APACrefbtitle {{MNIST} handwritten digit database, {Yann} {LeCun}, {Corinna}
  {Cortes} and {Chris} {Burges}.} {{MNIST} handwritten digit database, {Yann}
  {LeCun}, {Corinna} {Cortes} and {Chris} {Burges}.}
\newblock
\begin{APACrefURL}
  [{2021-07-30}]\url{http://yann.lecun.com/exdb/mnist/index.html}
  \end{APACrefURL}
\PrintBackRefs{\CurrentBib}

\bibitem [\protect \citeauthoryear {%
Long%
, Shelhamer%
\BCBL {}\ \BBA {} Darrell%
}{%
Long%
\ \protect \BOthers {.}}{%
{\protect \APACyear {2015}}%
}]{%
long_fully_2015}
\APACinsertmetastar {%
long_fully_2015}%
\begin{APACrefauthors}%
Long, J.%
, Shelhamer, E.%
\BCBL {}\ \BBA {} Darrell, T.%
\end{APACrefauthors}%
\unskip\
\newblock
\APACrefYearMonthDay{2015}{}{}.
\newblock
{\BBOQ}\APACrefatitle {Fully convolutional networks for semantic segmentation}
  {Fully convolutional networks for semantic segmentation}.{\BBCQ}
\newblock
\APACjournalVolNumPages{arXiv:1411.4038 [cs]}{}{}{}.
\PrintBackRefs{\CurrentBib}

\bibitem [\protect \citeauthoryear {%
Ma%
, Czerniawski%
\BCBL {}\ \BBA {} Leite%
}{%
Ma%
\ \protect \BOthers {.}}{%
{\protect \APACyear {2020}}%
}]{%
ma_semantic_2020}
\APACinsertmetastar {%
ma_semantic_2020}%
\begin{APACrefauthors}%
Ma, J\BPBI W.%
, Czerniawski, T.%
\BCBL {}\ \BBA {} Leite, F.%
\end{APACrefauthors}%
\unskip\
\newblock
\APACrefYearMonthDay{2020}{}{}.
\newblock
{\BBOQ}\APACrefatitle {Semantic segmentation of point clouds of building
  interiors with deep learning: {Augmenting} training datasets with synthetic
  {BIM}-based point clouds} {Semantic segmentation of point clouds of building
  interiors with deep learning: {Augmenting} training datasets with synthetic
  {BIM}-based point clouds}.{\BBCQ}
\newblock
\APACjournalVolNumPages{Automation in Construction}{113}{}{103144}.
\newblock
\begin{APACrefDOI} \doi{10.1016/j.autcon.2020.103144} \end{APACrefDOI}
\PrintBackRefs{\CurrentBib}

\bibitem [\protect \citeauthoryear {%
Mehta%
\ \BBA {} Sivaswamy%
}{%
Mehta%
\ \BBA {} Sivaswamy%
}{%
{\protect \APACyear {2017}}%
}]{%
mehta_m-net_2017}
\APACinsertmetastar {%
mehta_m-net_2017}%
\begin{APACrefauthors}%
Mehta, R.%
\BCBT {}\ \BBA {} Sivaswamy, J.%
\end{APACrefauthors}%
\unskip\
\newblock
\APACrefYearMonthDay{2017}{}{}.
\newblock
{\BBOQ}\APACrefatitle {M-net: {A} convolutional neural network for deep brain
  structure segmentation} {M-net: {A} convolutional neural network for deep
  brain structure segmentation}.{\BBCQ}
\newblock
\BIn{} \APACrefbtitle {2017 {IEEE} 14th {International} {Symposium} on
  {Biomedical} {Imaging} ({ISBI} 2017)} {2017 {IEEE} 14th {International}
  {Symposium} on {Biomedical} {Imaging} ({ISBI} 2017)}\ (\BPGS\ 437--440).
\newblock
\begin{APACrefDOI} \doi{10.1109/ISBI.2017.7950555} \end{APACrefDOI}
\PrintBackRefs{\CurrentBib}

\bibitem [\protect \citeauthoryear {%
Milletari%
, Navab%
\BCBL {}\ \BBA {} Ahmadi%
}{%
Milletari%
\ \protect \BOthers {.}}{%
{\protect \APACyear {2016}}%
}]{%
milletari_v-net_2016}
\APACinsertmetastar {%
milletari_v-net_2016}%
\begin{APACrefauthors}%
Milletari, F.%
, Navab, N.%
\BCBL {}\ \BBA {} Ahmadi, S\BHBI A.%
\end{APACrefauthors}%
\unskip\
\newblock
\APACrefYearMonthDay{2016}{}{}.
\newblock
{\BBOQ}\APACrefatitle {V-{Net}: {Fully} convolutional neural networks for
  volumetric medical image segmentation} {V-{Net}: {Fully} convolutional neural
  networks for volumetric medical image segmentation}.{\BBCQ}
\newblock
\APACjournalVolNumPages{arXiv:1606.04797 [cs]}{}{}{}.
\PrintBackRefs{\CurrentBib}

\bibitem [\protect \citeauthoryear {%
Mou%
, Hua%
\BCBL {}\ \BBA {} Zhu%
}{%
Mou%
\ \protect \BOthers {.}}{%
{\protect \APACyear {2019}}%
}]{%
mou_relation-augmented_2019}
\APACinsertmetastar {%
mou_relation-augmented_2019}%
\begin{APACrefauthors}%
Mou, L.%
, Hua, Y.%
\BCBL {}\ \BBA {} Zhu, X\BPBI X.%
\end{APACrefauthors}%
\unskip\
\newblock
\APACrefYearMonthDay{2019}{}{}.
\newblock
{\BBOQ}\APACrefatitle {A relation-augmented fully convolutional network for
  semantic segmentation in aerial scenes} {A relation-augmented fully
  convolutional network for semantic segmentation in aerial scenes}.{\BBCQ}
\newblock
\BIn{} (\BPGS\ 12416--12425).
\PrintBackRefs{\CurrentBib}

\bibitem [\protect \citeauthoryear {%
Ou%
\ \protect \BOthers {.}}{%
Ou%
\ \protect \BOthers {.}}{%
{\protect \APACyear {2019}}%
}]{%
ou_moving_2019}
\APACinsertmetastar {%
ou_moving_2019}%
\begin{APACrefauthors}%
Ou, X.%
, Yan, P.%
, Zhang, Y.%
, Tu, B.%
, Zhang, G.%
, Wu, J.%
\BCBL {}\ \BBA {} Li, W.%
\end{APACrefauthors}%
\unskip\
\newblock
\APACrefYearMonthDay{2019}{}{}.
\newblock
{\BBOQ}\APACrefatitle {Moving object detection method via {ResNet}-18 with
  encoder–decoder structure in complex scenes} {Moving object detection
  method via {ResNet}-18 with encoder–decoder structure in complex
  scenes}.{\BBCQ}
\newblock
\APACjournalVolNumPages{IEEE Access}{7}{}{108152--108160}.
\newblock
\begin{APACrefDOI} \doi{10.1109/ACCESS.2019.2931922} \end{APACrefDOI}
\PrintBackRefs{\CurrentBib}

\bibitem [\protect \citeauthoryear {%
Paszke%
\ \protect \BOthers {.}}{%
Paszke%
\ \protect \BOthers {.}}{%
{\protect \APACyear {2019}}%
}]{%
NEURIPS2019_9015}
\APACinsertmetastar {%
NEURIPS2019_9015}%
\begin{APACrefauthors}%
Paszke, A.%
, Gross, S.%
, Massa, F.%
, Lerer, A.%
, Bradbury, J.%
, Chanan, G.%
\BDBL {}Chintala, S.%
\end{APACrefauthors}%
\unskip\
\newblock
\APACrefYearMonthDay{2019}{}{}.
\newblock
{\BBOQ}\APACrefatitle {PyTorch: An imperative style, high-performance deep
  learning library} {Pytorch: An imperative style, high-performance deep
  learning library}.{\BBCQ}
\newblock
\BIn{} H.~Wallach, H.~Larochelle, A.~Beygelzimer, F.~d\textquotesingle
  Alch\'{e}-Buc, E.~Fox\BCBL {}\ \BBA {} R.~Garnett\ (\BEDS), \APACrefbtitle
  {Advances in Neural Information Processing Systems 32} {Advances in neural
  information processing systems 32}\ (\BPGS\ 8024--8035).
\newblock
\APACaddressPublisher{}{Curran Associates, Inc.}
\newblock
\begin{APACrefURL}
  \url{http://papers.neurips.cc/paper/9015-pytorch-an-imperative-style-high-performance-deep-learning-library.pdf}
  \end{APACrefURL}
\PrintBackRefs{\CurrentBib}

\bibitem [\protect \citeauthoryear {%
Ronneberger%
, Fischer%
\BCBL {}\ \BBA {} Brox%
}{%
Ronneberger%
\ \protect \BOthers {.}}{%
{\protect \APACyear {2015}}%
}]{%
ronneberger_u-net_2015}
\APACinsertmetastar {%
ronneberger_u-net_2015}%
\begin{APACrefauthors}%
Ronneberger, O.%
, Fischer, P.%
\BCBL {}\ \BBA {} Brox, T.%
\end{APACrefauthors}%
\unskip\
\newblock
\APACrefYearMonthDay{2015}{}{}.
\newblock
{\BBOQ}\APACrefatitle {U-{Net}: {Convolutional} networks for biomedical image
  segmentation} {U-{Net}: {Convolutional} networks for biomedical image
  segmentation}.{\BBCQ}
\newblock
\APACjournalVolNumPages{arXiv:1505.04597 [cs]}{}{}{}.
\PrintBackRefs{\CurrentBib}

\bibitem [\protect \citeauthoryear {%
Schlemper%
\ \protect \BOthers {.}}{%
Schlemper%
\ \protect \BOthers {.}}{%
{\protect \APACyear {2019}}%
}]{%
schlemper_attention_2019}
\APACinsertmetastar {%
schlemper_attention_2019}%
\begin{APACrefauthors}%
Schlemper, J.%
, Oktay, O.%
, Schaap, M.%
, Heinrich, M.%
, Kainz, B.%
, Glocker, B.%
\BCBL {}\ \BBA {} Rueckert, D.%
\end{APACrefauthors}%
\unskip\
\newblock
\APACrefYearMonthDay{2019}{}{}.
\newblock
{\BBOQ}\APACrefatitle {Attention gated networks: {Learning} to leverage salient
  regions in medical images} {Attention gated networks: {Learning} to leverage
  salient regions in medical images}.{\BBCQ}
\newblock
\APACjournalVolNumPages{Medical Image Analysis}{53}{}{197--207}.
\newblock
\begin{APACrefDOI} \doi{10.1016/j.media.2019.01.012} \end{APACrefDOI}
\PrintBackRefs{\CurrentBib}

\bibitem [\protect \citeauthoryear {%
Seo%
, Duque%
\BCBL {}\ \BBA {} Wacker%
}{%
Seo%
\ \protect \BOthers {.}}{%
{\protect \APACyear {2018}}%
}]{%
seo_drone-enabled_2018}
\APACinsertmetastar {%
seo_drone-enabled_2018}%
\begin{APACrefauthors}%
Seo, J.%
, Duque, L.%
\BCBL {}\ \BBA {} Wacker, J.%
\end{APACrefauthors}%
\unskip\
\newblock
\APACrefYearMonthDay{2018}{}{}.
\newblock
{\BBOQ}\APACrefatitle {Drone-enabled bridge inspection methodology and
  application} {Drone-enabled bridge inspection methodology and
  application}.{\BBCQ}
\newblock
\APACjournalVolNumPages{Automation in Construction}{94}{}{112--126}.
\newblock
\begin{APACrefDOI} \doi{10.1016/j.autcon.2018.06.006} \end{APACrefDOI}
\PrintBackRefs{\CurrentBib}

\bibitem [\protect \citeauthoryear {%
Shim%
\ \BBA {} Cho%
}{%
Shim%
\ \BBA {} Cho%
}{%
{\protect \APACyear {2020}}%
}]{%
shim_lightweight_2020}
\APACinsertmetastar {%
shim_lightweight_2020}%
\begin{APACrefauthors}%
Shim, S.%
\BCBT {}\ \BBA {} Cho, G\BHBI C.%
\end{APACrefauthors}%
\unskip\
\newblock
\APACrefYearMonthDay{2020}{}{}.
\newblock
{\BBOQ}\APACrefatitle {Lightweight semantic segmentation for road-surface
  damage recognition based on multiscale learning} {Lightweight semantic
  segmentation for road-surface damage recognition based on multiscale
  learning}.{\BBCQ}
\newblock
\APACjournalVolNumPages{IEEE Access}{8}{}{102680--102690}.
\newblock
\begin{APACrefDOI} \doi{10.1109/ACCESS.2020.2998427} \end{APACrefDOI}
\PrintBackRefs{\CurrentBib}

\bibitem [\protect \citeauthoryear {%
Simonyan%
\ \BBA {} Zisserman%
}{%
Simonyan%
\ \BBA {} Zisserman%
}{%
{\protect \APACyear {2015}}%
}]{%
simonyan_very_2015}
\APACinsertmetastar {%
simonyan_very_2015}%
\begin{APACrefauthors}%
Simonyan, K.%
\BCBT {}\ \BBA {} Zisserman, A.%
\end{APACrefauthors}%
\unskip\
\newblock
\APACrefYearMonthDay{2015}{}{}.
\newblock
{\BBOQ}\APACrefatitle {Very deep convolutional networks for large-scale image
  recognition} {Very deep convolutional networks for large-scale image
  recognition}.{\BBCQ}
\newblock
\APACjournalVolNumPages{arXiv:1409.1556 [cs]}{}{}{}.
\PrintBackRefs{\CurrentBib}

\bibitem [\protect \citeauthoryear {%
{Stanford Vision Lab}%
}{%
{Stanford Vision Lab}%
}{%
{\protect \APACyear {2020}}%
}]{%
stanford_vision_lab_imagenet_2020}
\APACinsertmetastar {%
stanford_vision_lab_imagenet_2020}%
\begin{APACrefauthors}%
{Stanford Vision Lab}.%
\end{APACrefauthors}%
\unskip\
\newblock
\APACrefYearMonthDay{2020}{}{}.
\newblock
\APACrefbtitle {{ImageNet}.} {{ImageNet}.}
\newblock
\begin{APACrefURL} [{2021-07-30}]\url{https://image-net.org/index.php}
  \end{APACrefURL}
\PrintBackRefs{\CurrentBib}

\bibitem [\protect \citeauthoryear {%
Tian%
\ \protect \BOthers {.}}{%
Tian%
\ \protect \BOthers {.}}{%
{\protect \APACyear {2020}}%
}]{%
tian_attention-guided_2020}
\APACinsertmetastar {%
tian_attention-guided_2020}%
\begin{APACrefauthors}%
Tian, C.%
, Xu, Y.%
, Li, Z.%
, Zuo, W.%
, Fei, L.%
\BCBL {}\ \BBA {} Liu, H.%
\end{APACrefauthors}%
\unskip\
\newblock
\APACrefYearMonthDay{2020}{}{}.
\newblock
{\BBOQ}\APACrefatitle {Attention-guided {CNN} for image denoising}
  {Attention-guided {CNN} for image denoising}.{\BBCQ}
\newblock
\APACjournalVolNumPages{Neural Networks}{124}{}{117--129}.
\newblock
\begin{APACrefDOI} \doi{10.1016/j.neunet.2019.12.024} \end{APACrefDOI}
\PrintBackRefs{\CurrentBib}

\bibitem [\protect \citeauthoryear {%
Tong%
, Xu%
\BCBL {}\ \BBA {} Denœux%
}{%
Tong%
\ \protect \BOthers {.}}{%
{\protect \APACyear {2021}}%
}]{%
tong_evidential_2021}
\APACinsertmetastar {%
tong_evidential_2021}%
\begin{APACrefauthors}%
Tong, Z.%
, Xu, P.%
\BCBL {}\ \BBA {} Denœux, T.%
\end{APACrefauthors}%
\unskip\
\newblock
\APACrefYearMonthDay{2021}{}{}.
\newblock
{\BBOQ}\APACrefatitle {Evidential fully convolutional network for semantic
  segmentation} {Evidential fully convolutional network for semantic
  segmentation}.{\BBCQ}
\newblock
\APACjournalVolNumPages{Applied Intelligence}{}{}{}.
\newblock
\begin{APACrefDOI} \doi{10.1007/s10489-021-02327-0} \end{APACrefDOI}
\PrintBackRefs{\CurrentBib}

\bibitem [\protect \citeauthoryear {%
{Torch Contributors}%
}{%
{Torch Contributors}%
}{%
{\protect \APACyear {2019}}%
}]{%
torch_contributors_bceloss_2019}
\APACinsertmetastar {%
torch_contributors_bceloss_2019}%
\begin{APACrefauthors}%
{Torch Contributors}.%
\end{APACrefauthors}%
\unskip\
\newblock
\APACrefYearMonthDay{2019}{}{}.
\newblock
\APACrefbtitle {{BCELoss} — {PyTorch} 1.9.1 documentation.} {{BCELoss} —
  {PyTorch} 1.9.1 documentation.}
\newblock
\begin{APACrefURL}
  [{2021-10-11}]\url{https://pytorch.org/docs/stable/generated/torch.nn.BCELoss.html}
  \end{APACrefURL}
\PrintBackRefs{\CurrentBib}

\bibitem [\protect \citeauthoryear {%
Usama%
\ \protect \BOthers {.}}{%
Usama%
\ \protect \BOthers {.}}{%
{\protect \APACyear {2020}}%
}]{%
usama_attention-based_2020}
\APACinsertmetastar {%
usama_attention-based_2020}%
\begin{APACrefauthors}%
Usama, M.%
, Ahmad, B.%
, Song, E.%
, Hossain, M\BPBI S.%
, Alrashoud, M.%
\BCBL {}\ \BBA {} Muhammad, G.%
\end{APACrefauthors}%
\unskip\
\newblock
\APACrefYearMonthDay{2020}{}{}.
\newblock
{\BBOQ}\APACrefatitle {Attention-based sentiment analysis using convolutional
  and recurrent neural network} {Attention-based sentiment analysis using
  convolutional and recurrent neural network}.{\BBCQ}
\newblock
\APACjournalVolNumPages{Future Generation Computer Systems}{113}{}{571--578}.
\newblock
\begin{APACrefDOI} \doi{10.1016/j.future.2020.07.022} \end{APACrefDOI}
\PrintBackRefs{\CurrentBib}

\bibitem [\protect \citeauthoryear {%
Xia%
\ \BBA {} Kulis%
}{%
Xia%
\ \BBA {} Kulis%
}{%
{\protect \APACyear {2017}}%
}]{%
xia_w-net_2017}
\APACinsertmetastar {%
xia_w-net_2017}%
\begin{APACrefauthors}%
Xia, X.%
\BCBT {}\ \BBA {} Kulis, B.%
\end{APACrefauthors}%
\unskip\
\newblock
\APACrefYearMonthDay{2017}{}{}.
\newblock
{\BBOQ}\APACrefatitle {W-{Net}: {A} deep model for fully unsupervised image
  segmentation} {W-{Net}: {A} deep model for fully unsupervised image
  segmentation}.{\BBCQ}
\newblock
\APACjournalVolNumPages{arXiv:1711.08506 [cs]}{}{}{}.
\PrintBackRefs{\CurrentBib}

\bibitem [\protect \citeauthoryear {%
Xiong%
, He%
, Liu%
\BCBL {}\ \BBA {} Liao%
}{%
Xiong%
\ \protect \BOthers {.}}{%
{\protect \APACyear {2020}}%
}]{%
xiong_end--end_2020}
\APACinsertmetastar {%
xiong_end--end_2020}%
\begin{APACrefauthors}%
Xiong, D.%
, He, C.%
, Liu, X.%
\BCBL {}\ \BBA {} Liao, M.%
\end{APACrefauthors}%
\unskip\
\newblock
\APACrefYearMonthDay{2020}{}{}.
\newblock
{\BBOQ}\APACrefatitle {An {end}-{to}-{end} {bayesian} {segmentation} {network}
  {based} on a {generative} {adversarial} {network} for {remote} {sensing}
  {images}} {An {end}-{to}-{end} {bayesian} {segmentation} {network} {based} on
  a {generative} {adversarial} {network} for {remote} {sensing}
  {images}}.{\BBCQ}
\newblock
\APACjournalVolNumPages{Remote Sensing}{12}{2}{216}.
\newblock
\begin{APACrefDOI} \doi{10.3390/rs12020216} \end{APACrefDOI}
\PrintBackRefs{\CurrentBib}

\bibitem [\protect \citeauthoryear {%
G.~Zhang%
, Pan%
\BCBL {}\ \BBA {} Zhang%
}{%
G.~Zhang%
\ \protect \BOthers {.}}{%
{\protect \APACyear {2021}}%
}]{%
zhang_semi-supervised_2021}
\APACinsertmetastar {%
zhang_semi-supervised_2021}%
\begin{APACrefauthors}%
Zhang, G.%
, Pan, Y.%
\BCBL {}\ \BBA {} Zhang, L.%
\end{APACrefauthors}%
\unskip\
\newblock
\APACrefYearMonthDay{2021}{}{}.
\newblock
{\BBOQ}\APACrefatitle {Semi-supervised learning with {GAN} for automatic defect
  detection from images} {Semi-supervised learning with {GAN} for automatic
  defect detection from images}.{\BBCQ}
\newblock
\APACjournalVolNumPages{Automation in Construction}{128}{}{103764}.
\newblock
\begin{APACrefDOI} \doi{10.1016/j.autcon.2021.103764} \end{APACrefDOI}
\PrintBackRefs{\CurrentBib}

\bibitem [\protect \citeauthoryear {%
Y.~Zhang%
, Miao%
, Mansi%
\BCBL {}\ \BBA {} Liao%
}{%
Y.~Zhang%
\ \protect \BOthers {.}}{%
{\protect \APACyear {2020}}%
}]{%
zhang_unsupervised_2020}
\APACinsertmetastar {%
zhang_unsupervised_2020}%
\begin{APACrefauthors}%
Zhang, Y.%
, Miao, S.%
, Mansi, T.%
\BCBL {}\ \BBA {} Liao, R.%
\end{APACrefauthors}%
\unskip\
\newblock
\APACrefYearMonthDay{2020}{}{}.
\newblock
{\BBOQ}\APACrefatitle {Unsupervised {X}-ray image segmentation with task driven
  generative adversarial networks} {Unsupervised {X}-ray image segmentation
  with task driven generative adversarial networks}.{\BBCQ}
\newblock
\APACjournalVolNumPages{Medical Image Analysis}{62}{}{101664}.
\newblock
\begin{APACrefDOI} \doi{10.1016/j.media.2020.101664} \end{APACrefDOI}
\PrintBackRefs{\CurrentBib}

\bibitem [\protect \citeauthoryear {%
Zhao%
, Shadabfar%
, Zhang%
, Chen%
\BCBL {}\ \BBA {} Huang%
}{%
Zhao%
\ \protect \BOthers {.}}{%
{\protect \APACyear {2021}}%
}]{%
zhao_deep_2021}
\APACinsertmetastar {%
zhao_deep_2021}%
\begin{APACrefauthors}%
Zhao, S.%
, Shadabfar, M.%
, Zhang, D.%
, Chen, J.%
\BCBL {}\ \BBA {} Huang, H.%
\end{APACrefauthors}%
\unskip\
\newblock
\APACrefYearMonthDay{2021}{}{}.
\newblock
{\BBOQ}\APACrefatitle {Deep learning-based classification and instance
  segmentation of leakage-area and scaling images of shield tunnel linings}
  {Deep learning-based classification and instance segmentation of leakage-area
  and scaling images of shield tunnel linings}.{\BBCQ}
\newblock
\APACjournalVolNumPages{Structural Control and Health
  Monitoring}{28}{6}{e2732}.
\newblock
\begin{APACrefDOI} \doi{10.1002/stc.2732} \end{APACrefDOI}
\PrintBackRefs{\CurrentBib}

\bibitem [\protect \citeauthoryear {%
Zheng%
\ \protect \BOthers {.}}{%
Zheng%
\ \protect \BOthers {.}}{%
{\protect \APACyear {2021}}%
}]{%
zheng_rethinking_2021}
\APACinsertmetastar {%
zheng_rethinking_2021}%
\begin{APACrefauthors}%
Zheng, S.%
, Lu, J.%
, Zhao, H.%
, Zhu, X.%
, Luo, Z.%
, Wang, Y.%
\BDBL {}Zhang, L.%
\end{APACrefauthors}%
\unskip\
\newblock
\APACrefYearMonthDay{2021}{}{}.
\newblock
{\BBOQ}\APACrefatitle {Rethinking {semantic} {segmentation} from a
  {sequence}-to-{sequence} {perspective} with {transformers}} {Rethinking
  {semantic} {segmentation} from a {sequence}-to-{sequence} {perspective} with
  {transformers}}.{\BBCQ}
\newblock
\BIn{} \APACrefbtitle {Proceedings of the {IEEE}/{CVF} {Conference} on
  {Computer} {Vision} and {Pattern} {Recognition} ({CVPR})} {Proceedings of the
  {IEEE}/{CVF} {Conference} on {Computer} {Vision} and {Pattern} {Recognition}
  ({CVPR})}\ (\BPGS\ 12196--12205).
\newblock
\begin{APACrefURL} [{2021-10-09}]\url{http://arxiv.org/abs/2012.15840}
  \end{APACrefURL}
\newblock
\APACrefnote{arXiv: 2012.15840}
\PrintBackRefs{\CurrentBib}

\bibitem [\protect \citeauthoryear {%
Zhou%
, Siddiquee%
, Tajbakhsh%
\BCBL {}\ \BBA {} Liang%
}{%
Zhou%
\ \protect \BOthers {.}}{%
{\protect \APACyear {2018}}%
}]{%
zhou_unet_2018}
\APACinsertmetastar {%
zhou_unet_2018}%
\begin{APACrefauthors}%
Zhou, Z.%
, Siddiquee, M\BPBI M\BPBI R.%
, Tajbakhsh, N.%
\BCBL {}\ \BBA {} Liang, J.%
\end{APACrefauthors}%
\unskip\
\newblock
\APACrefYearMonthDay{2018}{}{}.
\newblock
{\BBOQ}\APACrefatitle {{UNet}++: {A} nested {U}-net architecture for medical
  image segmentation} {{UNet}++: {A} nested {U}-net architecture for medical
  image segmentation}.{\BBCQ}
\newblock
\APACjournalVolNumPages{arXiv:1807.10165 [cs, eess, stat]}{}{}{}.
\PrintBackRefs{\CurrentBib}

\end{thebibliography}

\end{document}